\documentclass[10pt,twocolumn,letterpaper]{article}

\usepackage{geometry}
\usepackage{cvpr}
\usepackage{times}
\usepackage{epsfig}
\usepackage{graphicx}
\usepackage{amsmath}
\usepackage{amssymb}
\usepackage[ruled,vlined]{algorithm2e}
\usepackage{arydshln}
\usepackage[accsupp]{axessibility}  

\usepackage[pagebackref,breaklinks,colorlinks]{hyperref}

\cvprfinalcopy 
\def\cvprPaperID{7436} 

\begin{document}

\title{Sparse Object-level Supervision for Instance Segmentation \\ with Pixel Embeddings}

\author{
 Adrian Wolny \quad Qin Yu \quad Constantin Pape \quad Anna Kreshuk \vspace{.4em}\\
 European Molecular Biology Laboratory (EMBL), 
 Heidelberg, Germany
}

\maketitle

\begin{abstract}
Most state-of-the-art instance segmentation methods have to be trained on densely annotated images. While difficult in general, this requirement is especially daunting for biomedical images, where domain expertise is often required for annotation and no large public data collections are available for pre-training. We propose to address the dense annotation bottleneck by introducing a proposal-free segmentation approach based on non-spatial embeddings, which exploits the structure of the learned embedding space to extract individual instances in a differentiable way. The segmentation loss can then be applied directly to instances and the overall pipeline can be trained in a fully- or weakly supervised manner. We consider the challenging case of positive-unlabeled supervision, where a novel self-supervised consistency loss is introduced for the unlabeled parts of the training data. We evaluate the proposed method on 2D and 3D segmentation problems in different microscopy modalities as well as on the Cityscapes and CVPPP instance segmentation benchmarks, achieving state-of-the-art results on the latter. 
\let\thefootnote\relax\footnotetext{Correspondence to: \href{mailto:anna.kreshuk@embl.de}{\texttt{anna.kreshuk@embl.de}}}
\let\thefootnote\relax\footnotetext{Code: \href{https://github.com/kreshuklab/spoco}{\texttt{github.com/kreshuklab/spoco}}}
\end{abstract}

\section{Introduction}

Instance segmentation is one of the key problems addressed by computer vision. It is important for many application domains, from astronomy to scene understanding in robotics, forming the basis for the analysis of individual object appearance. Biological imaging provides a particularly large set of use cases for the instance segmentation task, with imaging modalities ranging from natural photographs for phenotyping to electron microscopy for detailed analysis of cellular ultrastructure. The segmentation task is often posed in crowded 3D environments or their 2D projections with multiple overlapping objects. Additional challenges -- compared to segmentation in natural images -- come from the lack of large, publicly accessible, annotated training datasets that could serve for general-purpose backbone training. Most microscopy segmentation networks are therefore trained from scratch, using annotations produced by domain experts in their limited time.  

Over the recent years, several weakly supervised segmentation approaches have been introduced to lighten the necessary annotation burden. For natural images, image-level labels can serve as a surprisingly strong supervision thanks to the popular image classification datasets which include images of individual objects and can be used for pre-training \cite{Cholakkal2019}. There are no such collections in microscopy (see also Fig.~\ref{fig:light_microscopy} for a typical instance segmentation problem example where image-level labels would be of no help). 
Semi-supervised instance segmentation methods  \cite{bellver2019budget,bellver2020mask,chen2020semi} can create pseudo-labels in the unlabeled parts of the dataset. However, these methods require (weak) annotation of \emph{all} the objects in at least a subset of images -- a major obstacle for microscopy datasets which often contain hundreds of tightly clustered objects, in 3D. 

\begin{figure}[!htbp]
\begin{center}
\includegraphics[width=1.0\linewidth]{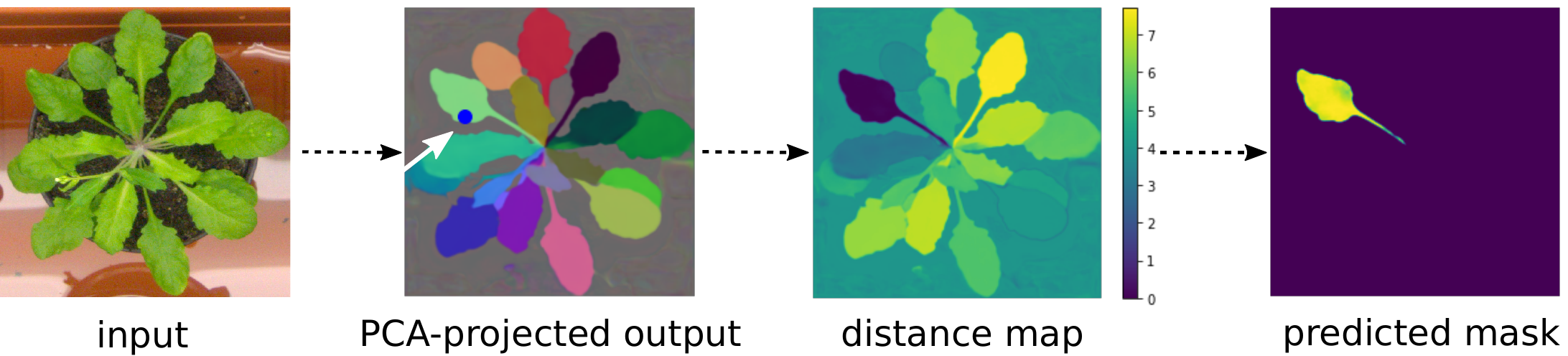}
\end{center}
\vspace{-0.8em}
   \caption{Differentiable instance selection for non-spatial embedding networks. First, we sample an anchor point randomly or guided by the groundtruth instances. Second, we compute a distance map in the embedding space from the anchor point to all image pixels. In the final step, a kernel function (Eq~\ref{eq:object}) transforms the distance map to the ``soft'' instance mask.}
\label{fig:inst_selection}
\end{figure}
\begin{figure*}[!htbp]
\begin{center}
\includegraphics[width=1.0\linewidth]{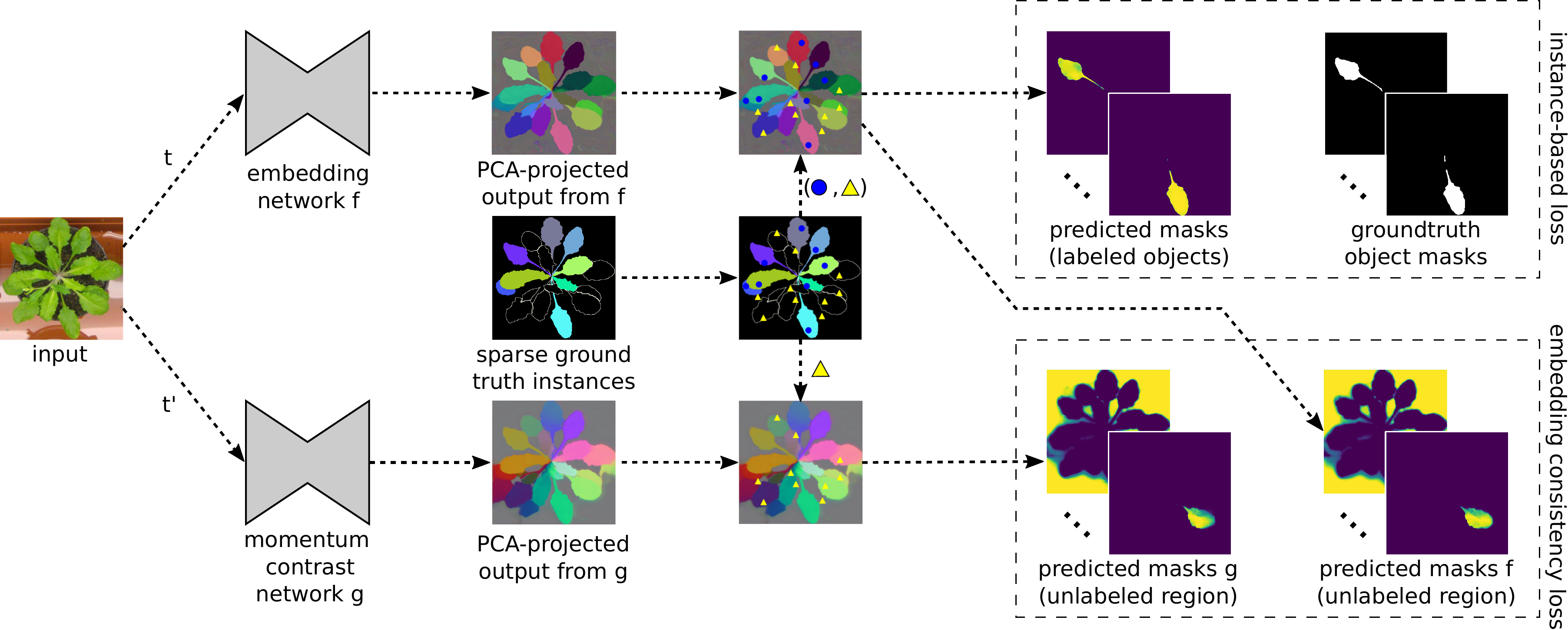}
\end{center}
\vspace{-0.8em}
   \caption{Overview of training procedure. Two augmented views of an input image are passed through two embedding networks $f(\cdot)$ and $g(\cdot)$ respectively. Anchor pixels inside labeled objects (blue dots) are sampled and their corresponding instances are extracted as shown in Fig.~\ref{fig:inst_selection}. Discrepancy between extracted objects and groundtruth objects is minimized by the instance-based loss. Another set of anchors (yellow triangles) is sampled exhaustively from the unlabeled region and for each anchor two instances are selected based on the outputs from $f(\cdot)$ and $g(\cdot)$. Discrepancy between instances is minimized using the embedding consistency loss.}
\vspace{-0.8em}
\label{fig:method}
\end{figure*}
The aim of our contribution is to address the dense annotation bottleneck by proposing a different kind of weak supervision for the instance segmentation problem: we require mask annotations for a subset of instances in the image, leaving the rest of the pixels unlabeled. This ``positive unlabeled'' setting has been explored in image classification and semantic segmentation problems \cite{PULearning2003, lejeune2021positiveunlabeled}, but - to the best of our knowledge - not for instance segmentation. Intrinsically, the instance segmentation problem is very well suited for positive unlabeled supervision:
as we also show empirically (Appendix A.5), sampling a few objects in each image instead of labeling a few images densely exposes the network to a more varied training set with better generalization potential. This is particularly important for datasets with sub-domains in the raw data distribution, as it can ensure all sub-domains are sampled without increasing the annotation time. Furthermore, in crowded microscopy images which commonly contain hundreds of objects, and especially in 3D, dense annotation is significantly more difficult and time consuming than sparse annotation, for the same total number of objects annotated. 
The main obstacle for training an instance segmentation method on sparse object mask annotations lies in the assignment of pixels to instances that happens in a non-differentiable step which precludes the loss from providing supervision at the level of individual instances. To lift this restriction, we propose a differentiable instance selection step which allows us to incorporate any (differentiable) \emph{instance-level} loss function into non-spatial pixel embedding network \cite{debrabandere2017semantic} training (Fig.~\ref{fig:inst_selection}). We show that with dense object mask annotations and thus full supervision, application of the loss at the single instance level consistently improves the segmentation accuracy of pixel embedding networks across a variety of datasets. For our main use case of weak positive unlabeled (PU) supervision, we propose to stabilize the training from sparse object masks by an additional instance-level consistency loss in the unlabeled areas of the images. The conceptually simple unlabeled consistency loss, inspired by \cite{He_2020_CVPR,tarvainen2017mean}, does not require the estimation of class prior distributions or the propagation of pseudo-labels, ubiquitously present in PU and other weakly supervised segmentation approaches \cite{Vernaza2017, Lin2016ScribbleSupSC}. In addition to training from scratch, our approach can deliver efficient domain adaptation using a few object masks in the target domain as supervision.

In summary, we address the instance segmentation task with a CNN that learns pixel embeddings and propose the first approach to enable training with weak positive unlabeled supervision, where only a subset of the object masks are annotated and no labels are given for the background. To this end, we introduce: (1) a differentiable instance selection step which allows to apply the loss directly to individual instances; (2) a consistency loss term that allows for instance-level training on unlabeled image regions, (3) a fast and scalable algorithm to convert the pixel embeddings into final instances, which partitions the metric graph derived from the embeddings.
We evaluate our approach on natural images (CVPPP \cite{MinerviniPRL2015} , Cityscapes \cite{Cityscapes_2016}) and microscopy datasets (2D and 3D, light and electron microscopy), reaching the state-of-the-art on CVPPP and consistently outperforming strong baselines for microscopy. On all datasets, the bulk of CNN performance improvement happens after just a fraction of training objects are annotated.  

\section{Related work}
Proposal-based methods such as Mask R-CNN \cite{he2017mask} are a popular choice for instance segmentation in natural images. These methods can be trained from weak bounding box labels \cite{khoreva2017simple,li2018weakly,song2019box,nishimura2019weakly}. However, as they require a pre-trained backbone network and have difficulties segmenting complex non-convex shapes, they have not become the go-to segmentation technique for microscopy imaging. There, instance segmentation methods commonly start from the semantic segmentation \cite{ronneberger2015u}, followed by a (non-differentiable) post-processing \cite{beier2017multicut,funke2017deep,lee2017superhuman,pape2019leveraging}.

Semantic instance segmentation with embedding networks was introduced by \cite{fathi2017semantic,debrabandere2017semantic}.
The embeddings of \cite{debrabandere2017semantic} have no explicit spatial or semantic component. \cite{fathi2017semantic} predicts a seediness score for each pixel in addition to the embedding vector.
The main advantage of pixel embedding-based segmentation methods lies in their superior performance for overlapping objects and crowded environments, delivering state-of-the-art results in many benchmarks, including those for biological data \cite{hirsch2020patchperpix}. Furthermore, they achieve a significant simplification of the pipeline: the same method can now be trained for intensity-based and for boundary-based segmentation. Our approach continues this line of work and employs non-spatial pixel embeddings.

Like the original proposal of \cite{debrabandere2017semantic}, all modern embedding networks require fully segmented images for training and compute the loss for the whole image rather than for the individual instances. Even when the supervision annotations are weak, such as scribbles or saliency masks, they are commonly used to create full object proposals or pseudo-labels to allow the loss to be applied to the whole image \cite{vangansbeke2021unsupervised, Vernaza2017, Lin2016ScribbleSupSC}. Such methods exploit object priors learned by their components which have been pre-trained on large public datasets. At the moment, such datasets or pre-trained backbones are not available for microscopy images. Another popular approach to weak supervision is to replace mask annotations by bounding boxes \cite{khoreva2017simple} which are much faster to produce. Given a pre-trained backbone, bounding boxes can be reduced to single point annotations \cite{Laradji2020}, but for training every object must be annotated, however weakly. The aim of our work is to lift this requirement and enable instance segmentation training with positive unlabeled supervision. 

Positive unlabeled learning targets classification problems where negative labels are unavailable or unreliable \cite{liu2003building, bekker2020learning}. Three approaches are in common use: generation of negative pseudo-labels, biased learning with class label noise in unlabeled areas and class prior incorporation (see \cite{bekker2020learning} for detailed review). PU learning has recently been extended to object detection \cite{bepler2019positive} and semantic segmentation problems \cite{lejeune2021positive}. Our approach enables PU learning for instance segmentation problems via an instance-level consistency loss applied to the unlabeled areas.  

The core of our approach consists of the differentiable single instance selection step performed during training. Here, we have drawn inspiration from \cite{Neven_2019_CVPR}, where the clustering bandwidth is learned in the network training which allows to optimize the intersection-over-union loss for each instance. Still, as  the network also needs to be trained to predict a seed map of cluster centers for inference, this method cannot be trained on partially labeled images. Differentiable single instance selection has also been proposed by AdaptIS \cite{sofiiuk2019adaptis}. However, this method does not use a learned pixel embedding space and thus requires an additional sub-network to perform instance selection. Importantly, AdaptIS does not introduce PU training and relies on a pre-trained backbone network which is not readily available for microscopy images. 

\section{Methods}

\subsection{Full supervision}
\label{ssec:single_obj}
Given an image $I=\{I_1, ..., I_C\}$ composed of $C$ objects (including background), $N_k$ pixels in $I_k$, $N = \sum_{k=1}^{C} N_{k}$ pixels in the image and an embedding network $f \colon \mathbb{R}^3 \to \mathbb{R}^D$ which maps pixel $i$ into a $D$-dimensional embedding vector $\boldsymbol{e}_i$, the discriminative loss \cite{debrabandere2017semantic} is defined by the \textit{pull force} and the \textit{push force} terms\footnote{Similarly to \cite{debrabandere2017semantic} a regularization term ($\frac{1}{C} \sum_{k=1}^{C} \lVert \mu_{k} \rVert$) which keeps the embeddings bounded is added to the final loss with a small weight of 0.001. For clarity, we omit this term in the text}:
\begin{align}
L_{pull} &= \frac{1}{C} \sum_{k=1}^{C} \frac{1}{N_{k}} \sum_{i=1}^{N_{k}} \left[ \lVert \boldsymbol{\mu}_k - \boldsymbol{e}_i \rVert - \delta_v \right]_{+}^2 \label{eq:variance_term}\\
L_{push} &= \frac{1}{C^2}\sum_{k=1}^{C} \sum_{l=1}^{C} [2\delta_d - \left\| \boldsymbol{\mu}_k - \boldsymbol{\mu}_l \right\|]_{+}^2 \label{eq:dist_term}
\end{align}
where $\lVert \cdot \rVert$ is the L2-norm and $[x]_{+}=max(0,x)$ is the rectifier function.
The pull force $L_{pull}$ (Eq.~\ref{eq:variance_term}) brings the object's pixel embeddings closer to their mean embedding $\boldsymbol{\mu}_k$, while the push force $L_{push}$ (Eq.~\ref{eq:dist_term}) pushes the objects away, by increasing the distance between mean instance embeddings. Note that both terms are hinged, i.e. embeddings within the $\delta_v$-neighbourhood of the mean embedding $\boldsymbol{\mu}_k$ are no longer pulled to it. Similarly, mean embeddings which are further apart than $2\delta_d$ are no longer repulsed.

We exploit the clustering induced by this loss to select pixels belonging to a single instance and apply auxiliary losses at the instance level (Fig.~\ref{fig:method}). 
Crucially, we find that given an instance $I_k$ it is possible to extract a ``soft'' mask $S_k$ for the current network prediction of the instance $I_k$ in a \emph{differentiable} way (Fig.~\ref{fig:inst_selection}). We select an anchor point for $I_k$ at random and project it into the learned embedding space to recover its embedding $\boldsymbol{a}_k$, which we term anchor embedding. We compute the distance map from all image pixel embeddings to the anchor embedding and apply a Gaussian kernel function $\phi \colon \mathbb{R}^D \times \mathbb{R}^D \to \mathbb{R}$ to ``softly'' select the pixels within the $\delta_v$-neighborhood of $\boldsymbol{a}_k$ ($\delta_v$ is the pull term margin in Eq.~\ref{eq:variance_term}): 
\begin{equation} 
\label{eq:object}
\begin{aligned}
S_k &= \{\phi(\boldsymbol{e}_i, \boldsymbol{a}_k) \mid i=1,...,N \}\\
\phi(\boldsymbol{e}_i, \boldsymbol{a}_k) &= \exp\left({ - \frac{ \| \boldsymbol{e}_i - \boldsymbol{a}_k \| ^ 2}{ 2 \sigma ^ 2}  }\right)
\end{aligned}
\end{equation}
We require the embeddings within the distance $\delta_v$ from the anchor embedding $\boldsymbol{a}_k$ have a kernel value greater than a predefined threshold $t \in (0, 1)$, i.e. $\phi(\boldsymbol{e}_i, \boldsymbol{a}_k) \geq t \iff \| \boldsymbol{e}_i - \boldsymbol{a}_k \| \leq \delta_v$. We can thus determine $\sigma^2$: substituting $\| \boldsymbol{e}_i - \boldsymbol{a}_k \| = \delta_v$ in Eq.~\ref{eq:object}, we get
$\exp\left({ - \frac{ \delta_v ^ 2}{ 2 \sigma ^ 2}  }\right) = t$, i.e. $\sigma^2 = \frac{-\delta_v^2}{2\ln t}$. We choose $t=0.9$ in our experiments and refer to Appendix A.7 for a detailed exploration of this hyperparameter.

We can now formulate a loss for a single object segmentation mask by minimizing the Dice loss ($D$) \cite{milletari2016vnet} of the mask $S_k$ predicted using Eq.~\ref{eq:object} and the corresponding groundtruth mask $I_k$. 
\begin{equation} \label{eq:object_loss_dice}
L_{obj} = \frac{1}{C}\sum_{k=1}^{C} D(S_k, I_k)
\end{equation}
Combining the losses in Eq.~\ref{eq:variance_term}, Eq.~\ref{eq:dist_term} and Eq.~\ref{eq:object_loss_dice}, we get:
\begin{equation}
\label{eq:cl_dice}
L_{SO} = \alpha L_{pull} + \beta L_{push} + \lambda L_{obj}
\end{equation}
which we refer to as the Single Object contrastive loss ($L_{SO}$). We use $\alpha = \beta = 1$ (similar to \cite{debrabandere2017semantic}) and $\lambda = 1$ in our experiments. We set the pull and push margin parameters to $\delta_v=0.5, \delta_d=2.0$.

While Eq.~\ref{eq:object_loss_dice} employs the Dice loss, our approach is not limited to Dice and can be used with any differentiable loss function at the single instance level, e.g. binary cross-entropy. Additionally, we explored the adversarial approach and trained a discriminator to distinguish the object masks coming from our differentiable instance selection method or from the groundtruth. More details can be found in Appendix A.1, the results are shown in Table~\ref{tab:ovules}.

 \subsection{Weak supervision}
 \label{ssec:sparse_single_obj}
 To enable training from positive unlabeled supervision, we introduce two additional loss terms: one to push each cluster away from the pixels in the unlabeled region and the other to enforce embedding space consistency in the unlabeled region. For an unlabeled region $U$ which can contain both background and unlabeled instances, we define an additional ``push'' term:
\begin{equation} \label{eq:ul_push}  
L_{U\_push} = \frac{1}{C}\sum_{k=1}^{C}\frac{1}{N_U}\sum_{i=1}^{N_U}[\delta_d - \left\| \boldsymbol{\mu}_k - \boldsymbol{e}_i \right\|]_{+}^2
\end{equation}
where $C$ is the number of labeled clusters/instances and $N_U$ is the number of pixels in the unlabeled region $U$.

\begin{figure}[!htbp]

\begin{center}
\includegraphics[width=1.0\linewidth]{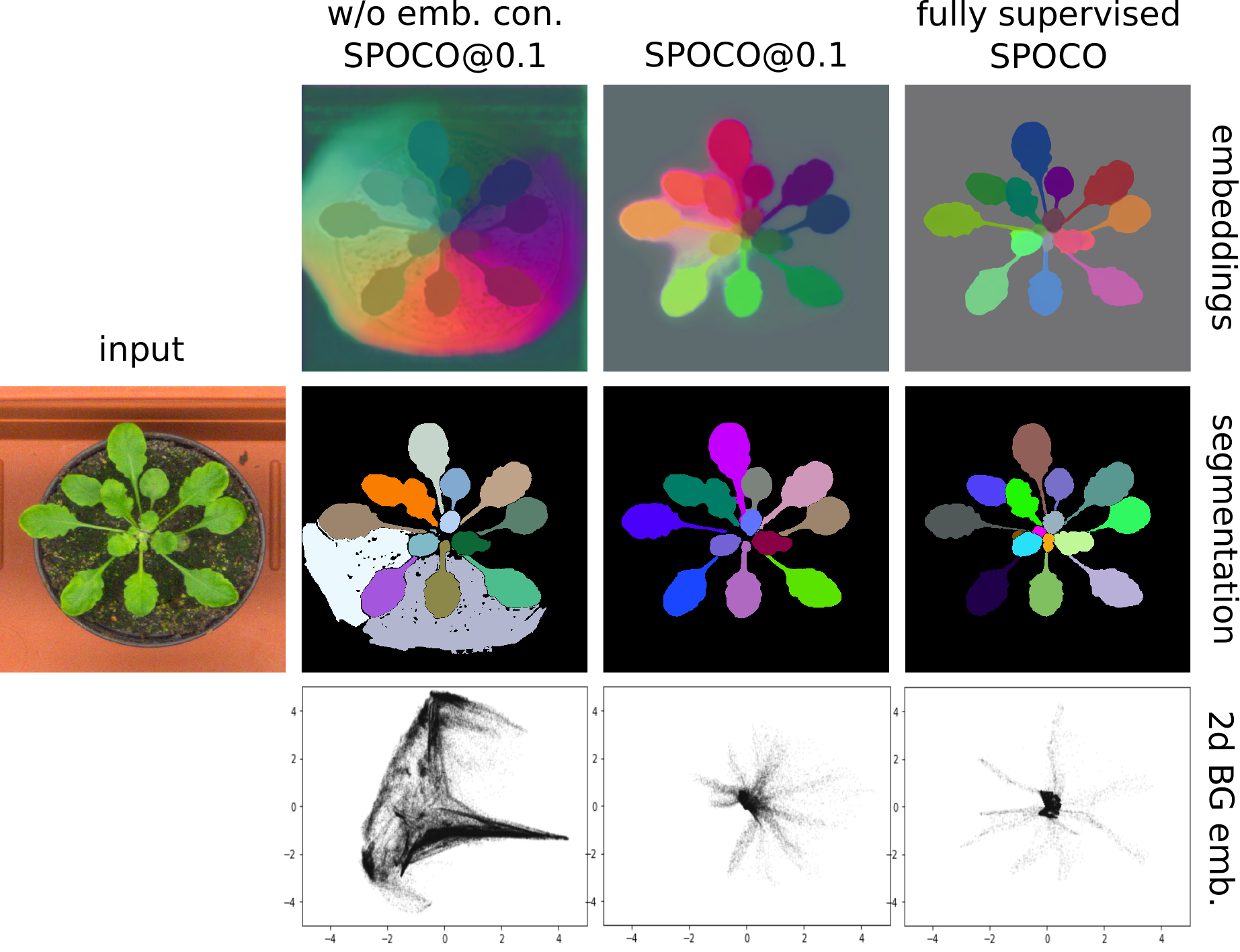}
\end{center}
\vspace{-0.8em}
   \caption{Different training schemes, left to right: SPOCO@0.1 trained without embedding consistency; SPOCO@0.1 trained with embedding consistency; SPOCO trained with full supervision (including the background label). \textbf{\textsc{Top)}} PCA-projected embeddings; \textbf{\textsc{Middle)}} corresponding clustering results; \textbf{\textsc{Bottom)}}  background pixel embeddings PCA-projected onto 2D subspace.}
\vspace{-0.8em}
\label{fig:embeddings}

\end{figure}

Since there is no direct supervision applied onto the unlabeled part of the image, the fully convolutional embedding network propagates the high frequency patterns present in there into the feature space. This is especially apparent for natural images and microscopy images with complex background structures, e.g. electron microscopy (see Fig~\ref{fig:embeddings} top left and Fig~\ref{fig:mitoem} top, col 3). To overcome this issue, we introduce the embedding consistency term. Given two different embedding networks $f$ and $g$, we perturb the input image $x$ with two different random, location- and shape-preserving augmentations $t$ and $t'$ and pass it through $f$ and $g$ respectively. The resulting vector fields $f(t(x))$ and $g(t'(x))$ come from the same input geometry, hence they should result in consistent instance segmentation after clustering, also in the unlabeled part of the input. To enforce this consistency we randomly sample an anchor point from the unlabeled region, project it into the $f$- and $g$-embedding spaces, to get anchor embeddings $\boldsymbol{a}^f$ and $\boldsymbol{a}^g$ and compute two ``soft'' masks $S^f$ and $S^g$ according to Eq.~\ref{eq:object}.
Similarly to Eq.~\ref{eq:object_loss_dice} the embedding consistency is given by maximising the overlap of the two masks, using the Dice loss ($D$):
\begin{equation} \label{eq:ul_con}  
L_{U\_con} = \frac{1}{K}\sum_{k=1}^{K} D(S^f_k, S^{g}_k)
\end{equation}
where $K$ is the number of anchor points sampled from the unlabeled region $U$ such that the whole region is covered by the union of extracted masks, i.e. $U \approx \bigcup_{k=1}^K {S^f_k \cup S^g_k}$. Having considered different variants of $g$-network including: weight sharing (with and without dropout) and independent training, we choose a momentum-based scheme \cite{He_2020_CVPR, NEURIPS2020_BYOL} where the network $g$ (parameterized by $\theta_g$) is implemented as an exponential moving average of the network $f$ (parameterized by $\theta_f$). The update rule for $\theta_g$ is given by: $\theta_g \leftarrow m \theta_g + (1 - m) \theta_f$. $f$ is trained by back-propagation. We refer to Appendix A.3 for extensive ablations of the $g$-network types and A.6 for the choice of a momentum coefficient $m\in [0, 1)$. Briefly, momentum variant provides the fastest convergence rate, improves training stability and is motivated by prior work \cite{tarvainen2017mean, dino2021}. Significance of the embedding consistency term in weakly supervised setting is illustrated in Fig.~\ref{fig:embeddings}. Note how the complex patterns present in the background (e.g. the flower pot) are propagated into the embedding space of the network trained without the consistency term (top, column 2), leading to spurious objects in the background after clustering (middle, column 2). In contrast, the same network trained with the embedding consistency loss results in crisp embeddings, homogeneous background embedding and clear background separation with no false positives (column 3).
We confirm this observation by PCA-projecting the embeddings of background pixels onto 2D subspace (bottom). Network trained sparsely with the consistency term implicitly pulls background pixels into a single cluster, similar to the fully supervised network where the background pull is enforced by the loss. In contrast, the network trained without the consistency loss does not form a tight background cluster in the feature space. 
In addition, with a limited annotation budget of a certain number of objects, we achieve (see Appendix A.5) much better segmentation accuracy with objects distributed across many images than with a few images fully labeled. The latter is prone to over-fitting, whereas a more diverse training set and the presence of the strong consistency regularizer in the former enables it to train from just a few object mask annotations. 
Our weakly supervised loss, termed Sparse Single Object loss ($L_{SSO}$), is given by:
\begin{equation}
\label{eq:socl}
L_{SSO} = \hat{L}_{SO} + \gamma \cdot L_{U\_push} + \delta \cdot L_{U\_con}
\end{equation}
In our experiments we use $\gamma = \delta = 1$. \\
Tab~\ref{tab:cityscapes} and Appendix A.2 shows that using the consistency term (Eq.~\ref{eq:ul_con}) in a fully-supervised setting, in addition to the instance-based term (Eq.~\ref{eq:object_loss_dice}) improves the segmentation accuracy at the expense of longer training times.
Fig.~\ref{fig:method} gives a graphical overview of our training procedure which we term \textit{SPOCO} (SParse Object COnsistency loss). Extensive ablation study of the individual loss terms can be found in Appendix A.2. In the experiments, we use the term SPOCO to refer to the fully-supervised training (taking all groundtruth objects including the background object). SPOCO@p refers to the weakly supervised positive unlabeled setting, in which a fraction $p \in (0, 1]$ of objects (excluding background) is taken for training. The background label is never selected in the weakly supervised setting, i.e. SPOCO@1.0 means that the network was trained with all labeled objects, excluding background.

\subsection{Clustering}
The final instance segmentation is obtained by clustering the predicted pixel embeddings. Mean-shift \cite{comaniciu2002mean} and HDBSCAN \cite{campello2013density} are commonly used for this task \cite{debrabandere2017semantic, kulikov2020instance, Payer2019b}. 
In this work we experimented with two additional clustering schemes: (1) partitioning \cite{mws2018} of a metric graph derived from pixel embeddings \cite{Lee2021} and (2) a hybrid approach where initial mean-shift clusters are refined to conform with the pull-push loss formulation (Sec~\ref{ssec:single_obj}). Embeddings from networks $f$ and $g$ are used together in (2), all other clustering methods use the $f$-embeddings only.
The advantage of (1) is a much faster inference time, whereas (2) can result in higher-quality segmentations. We refer to Appendix A.4 for a detailed description and comparison of different clustering methods.

\section{Experiments}
The fully- and semi-supervised evaluation is based on:

\textbf{CVPPP.} 
We use the A1 subset of the popular CVPPP dataset \cite{MinerviniPRL2015} which is part of the LSC competition. The task is to segment individual leaf instances of a plant growing in a pot. 
The dataset consists of 128 training images with public groundtruth labels and 33 test images with no publicly available labels. Test images come with a foreground mask which can be used during inference.

\textbf{Cityscapes.} 
We use Cityscapes \cite{Cityscapes_2016} to demonstrate the performance of our method on a large-scale instance-level segmentation of urban street scenes. There are 2975 training images, 500 validation images, and 1525 test images with fine annotations. We choose 8 semantic classes: \textit{person, rider, car, truck, bus, train, motorcycle, bicycle} and train the embedding networks separately for each class using the training set in the full and weak supervision setting. 

\textbf{Light microscopy (LM) datasets.} 
To evaluate the performance of our approach on a challenging boundary-based segmentation task we selected a 3D LM dataset of the ovules of \emph{Arabidopsis thaliana} from \cite{plantseg2020}, with 48 image stacks in total: 39 for training, 2 for validation and 7 for testing.
Additionally, we use the 3D \emph{A. thaliana} apical stem cells from \cite{WillisE8238pnas} in a transfer learning setting. The images are from the same imaging modality as the ovules dataset (confocal, cell membrane stained), but differ in cell type and image acquisition settings. We choose the Ovules dataset as the source domain and Stem cells as the target (\textit{plant1, plant2, plant4, plant13, plant15} are used for fine-tuning and \textit{plant18} for testing).

\textbf{Electron microscopy (EM) datasets.} 
Here we test our method in the transfer learning setting on the problem of mitochondria segmentation. An important difference between light and electron microscopy from the segmentation perspective lies in the appearance of the background which is simply dark and noisy for LM and highly structured for EM. The source domain (VNC dataset) \cite{zora91121} is a small annotated $20 \times 1024 \times 1024$px volume of the Drosophila larva VNC acquired with voxel size of $50 \times 5 \times 5$nm. We use 13 consecutive slices for training and keep 7 slices for validation. As target domain we use the 3D MitoEM-R dataset from the MitoEM Challenge \cite{wei2020mitoem} a $500 \times 4096 \times 4096$px volume at $30 \times 8 \times 8$ nm resolution extracted from rat cortex. Slices (0-399) are used for fine-tuning and (400-499) for testing.

\subsection{Setups}
\label{ssec:setups}
Any fully convolutional architecture with dense outputs could be used as an embedding network. We choose the U-Net \cite{ronneberger2015u, 3dunetCicekALBR16}. The depth of the U-Net is chosen such that the receptive field of features in the bottleneck layer is greater or equal to the input patch size. In all experiments we train the networks from scratch without using any per-trained backbones.
We use the Adam \cite{kingma2014adam} optimizer with initial learning rate $0.0002$ and weight decay $0.00001$. Data augmentation consists of random crops, random flips, random scaling and random elastic deformations. For the momentum contrast embedding network, we additionally use additive Gaussian noise, Gaussian blur and color jitter as geometry preserving transformations. 

In transfer learning experiments, the source network is always trained with the full groundtruth. On the target domain, we reduce the learning rate by a factor of 10 compared to the source network and use only a small fraction of the objects. VNC dataset is too small to train a 3D U-Net, so we perform EM segmentation in 2D, slice-by-slice. We also downsample VNC dataset by factor 1.6 in XY to match the voxel size of the target MitoEM data.

A detailed description of the network architecture, training procedure and hyperparameter selection can be found in the Appendix A.1.

\subsection{Results and Discussion}
\label{ssec:results}

\textbf{CVPPP.} 
Table~\ref{tab:cvppp} shows the results on the test set. The challenge provides foreground masks for test set images and  we assume they have been used by authors of \cite{debrabandere2017semantic, ren2017endtoend, kulikov2020instance} in test time inference. In this setting, SPOCO outperforms \cite{kulikov2020instance} and the current winner of the leaderboard on the A1 dataset, keeping the advantage even in the case when the foreground mask is not given, but learned by another network (``predicted FG''). Even without using the foreground mask in the final clustering, SPOCO is close to \cite{kulikov2020instance} in segmentation accuracy, achieving much better average difference in counting score ($|DiC|$).
We evaluate weakly supervised predictions without the foreground mask as we cannot easily train a semantic network without background labels. Nevertheless, even when training with only 10\% of the groundtruth instances (SPOCO@0.1), the Symmetric Best Dice ($SBD$) as compared with the fully supervised SPOCO (without FG) drops only by 10 percent points. Qualitative results from SPOCO@0.1 can be seen in Fig.~\ref{fig:embeddings} (column 3), where the single under-segmentation error is present in the top left part of the image. HDBSCAN with $min\_size=200$ is used for clustering in this case. Visual results and performance metrics for other clustering methods can be found in Appendix A.4. CVPPP dataset was used extensively in the ablation study, see A.2 for details.

\begin{table}
\begin{center}
\begin{tabularx}{\columnwidth}{|>{\hsize=1.0\hsize\linewidth=\hsize}X|>{\hsize=0.6\hsize\linewidth=\hsize}X|>{\hsize=0.4\hsize\linewidth=\hsize}X|}
\hline
Method & SBD & $|DiC|$ \\
\hline
Discriminative loss \cite{debrabandere2017semantic} & 0.842 & 1.0 \\
Recurrent attention \cite{ren2017endtoend} & 0.849 & \textbf{0.8} \\
Harmonic Emb. \cite{kulikov2020instance} & 0.899 & 3.0 \\
\hline
SPOCO (GT FG) & \textbf{0.932} & 1.7 \\
SPOCO (pred FG) & 0.920 & 1.6 \\
SPOCO (w/o FG) & 0.886 & 1.3 \\
\hline
SPOCO@0.1 & 0.788 $\pm$ 0.017 & 5.4 $\pm$ 0.3 \\
SPOCO@0.4 & 0.824 $\pm$ 0.003 & 3.2 $\pm$ 0.5 \\
SPOCO@0.8 & 0.828 $\pm$ 0.010 & 1.6 $\pm$ 0.2 \\
\hline
\end{tabularx}
\end{center}
\caption{Results on the CVPPP test set. Segmentation ($SBD$) and counting ($|DiC|$) scores for fully supervised SPOCO are reported in 3 different clustering settings: (1) with the groundtruth foreground mask, (2) with the predicted foreground mask (3) without the foreground mask. Results for semi-supervised setting SPOCO@p (no foreground mask) are presented for 10\%, 40\% and 80\% of randomly selected groundtruth instances.}
\vspace{-1em}
\label{tab:cvppp}
\end{table}

\begin{figure}[!htbp]
\begin{center}
\includegraphics[width=1.0\linewidth]{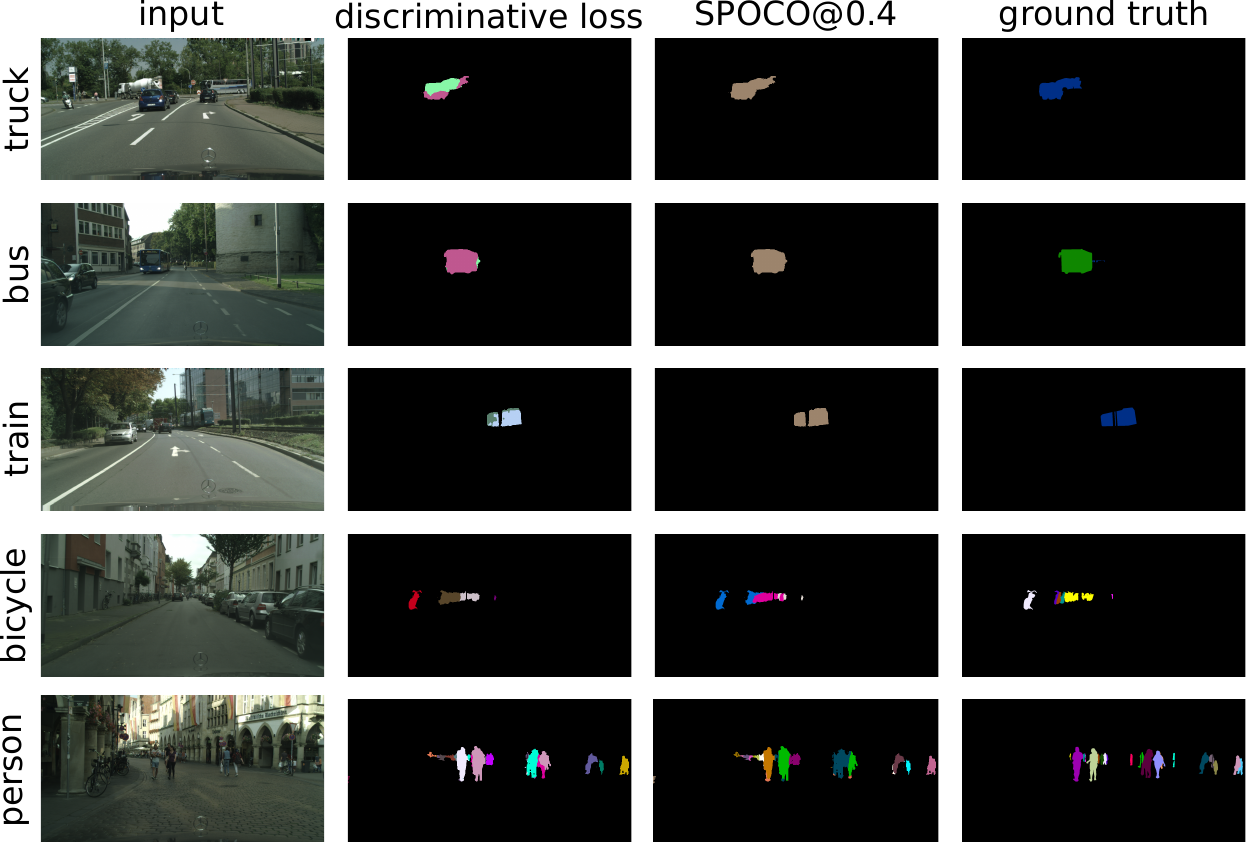}
\end{center}
\vspace{-0.8em}
   \caption{Segmentation results for randomly selected images of different semantic classes on the Cityscapes validation set.}
\label{fig:cityscapes} 
\end{figure}

\textbf{Cityscapes.}
We train our method with sparse (SPOCO@0.4) and full supervision and compare it with the fully-supervised contrastive framework \cite{debrabandere2017semantic}. In \cite{debrabandere2017semantic} authors trained a single model with multiple classes, applying the loss only within a given semantic mask. Since groundtruth semantic masks are not available when training from sparsely labeled instances, we train one model (including our implementation of \cite{debrabandere2017semantic}) for each semantic class. For inference we use pre-trained semantic segmentation model (DeepLabV3 \cite{deeplabv3_2018}) to generate semantic masks and cluster the embeddings only within a given semantic mask. After initial mean-shift clustering we merge every pair of clusters if the mean cluster embeddings are closer than $\delta_d$ (push force hinge in Eq.~\ref{eq:dist_term}).
Average Precision at 0.5 intersection-over-union computed on the validation set can be found in Table~\ref{tab:cityscapes}. Our method outperforms \cite{debrabandere2017semantic} with only 40\% of the groundtruth objects of each semantic class used for training. This is true for all classes apart from person, car and bicycle where the model requires larger number of annotated objects to reach high precision.
Importantly, using consistency term in the fully-supervised setting improves the score by a large margin. The performance of SPOCO@0.4 is almost as good as the fully-supervised SPOCO without the consistency term. We hypothesize that strong regularization induced by the consistency term is crucial for classes with small number of instances.
Fig.~\ref{fig:cityscapes} shows qualitative results on a few samples from the validation set. Network trained with discriminative loss frequently over-segments large instances (trucks, buses, trains). A common mistake in crowded scenes for both methods is the merging of neighboring instances.
Segmentation scores at different sampling rates, comparison with a class-agnostic training setting as well as qualitative results can be found in the Appendix A.8.

\begin{table}
\begin{center}
\begin{tabular}{|l|l|l|l|l|}
\hline
Class & DL \cite{debrabandere2017semantic} & S@0.4  & S w/ con & S w/o con \\
\hline
person              & \textbf{0.275} & 0.230 & 0.260          & 0.270 \\
rider               & 0.392          & 0.396 & \textbf{0.451} & 0.448 \\
car                 & \textbf{0.416} & 0.301 & 0.331          & 0.363 \\
truck               & 0.486          & 0.558 & \textbf{0.604} & 0.527 \\
bus                 & 0.504          & 0.601 & \textbf{0.637} & 0.530 \\
train               & 0.375          & 0.594 & \textbf{0.656} & 0.490 \\
motorcycle          & 0.382          & 0.405 & \textbf{0.464} & 0.461 \\
bicycle             & \textbf{0.267} & 0.214 & 0.266          & 0.255 \\
\hline
\textbf{average}    & 0.387 & 0.412 & \textbf{0.459}  & 0.418 \\
\hline
\end{tabular}
\end{center}
\caption{Segmentation results on the Cityscapes validation set. Average and per-class AP@0.5 scores are reported. \textit{DL} - discriminative loss \cite{debrabandere2017semantic}, \textit{S@0.4} - SPOCO@0.4, \textit{S w/ con} - fully-supervised SPOCO with the consistency term, \textit{S w/o con} - fully-supervised SPOCO without the consistency term.}
\vspace{-0.8em}
\label{tab:cityscapes}
\end{table}

\textbf{3D LM datasets.} 
We compare SPOCO to the method of \cite{plantseg2020}: a 3-step pipeline of boundary prediction, supervoxel generation and graph agglomeration. Following \cite{plantseg2020}, Adapted Rand Error \cite{CREMI2017} is used for evaluating the segmentation accuracy. As shown in Table~\ref{tab:ovules}, the performance of SPOCO is close to that of the much more complex 3-step PlantSeg pipeline. An additional adversarial loss term (SPOCO with $L_{wgan}$, see Appendix A.1) brings another performance boost and improves SPOCO accuracy beyond the \cite{plantseg2020} level. 

\begin{figure}[!htbp]
\begin{center}
\includegraphics[width=1.0\linewidth]{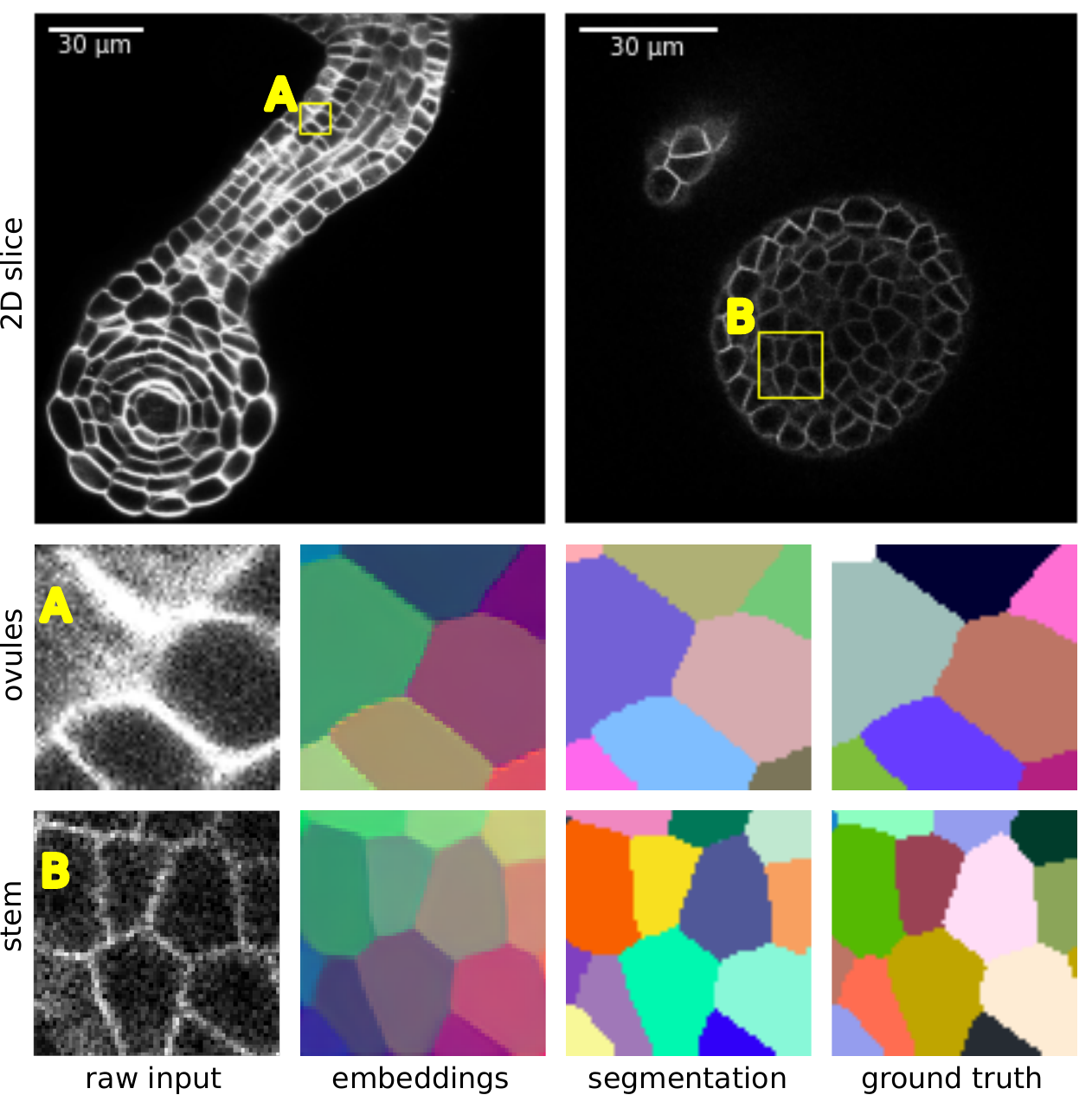}
\end{center}
\vspace{-0.5em}
   \caption{LM segmentation in standard and transfer learning settings. \textbf{\textsc{Top)}} samples from the 3D Ovules (left) and Stem (right) datasets; \textbf{\textsc{Middle)}} segmentation of a selected patch (A) from the source domain; \textbf{\textsc{Bottom)}} output of the source (Ovules) network fine-tuned with 1\% of instances from the target (Stem)  and the corresponding segmentation of a selected patch (B).}
\vspace{-0.5em}
\label{fig:light_microscopy} 
\end{figure}

Note that SPOCO trained with 10\% of the groundtruth instances already outperforms the original embedding network with discriminative loss \cite{debrabandere2017semantic}.  See Fig.~\ref{fig:light_microscopy} (top row) for qualitative results on a randomly sampled test set patch.

\begin{table}
\begin{center}
\begin{tabularx}{0.8\columnwidth}{|>{\hsize=1.3\hsize\linewidth=\hsize}X|>{\hsize=0.7\hsize\linewidth=\hsize}X|}
\hline
Method & ARand error \\
\hline
PlantSeg \cite{plantseg2020} & 0.046 \\
Discriminative loss \cite{debrabandere2017semantic} & 0.074 \\
\hline
SPOCO & 0.048 \\
SPOCO with $L_{wgan}$ & \textbf{0.042} \\
\hline
SPOCO@0.1 & 0.069  \\
SPOCO@0.4 & 0.060  \\
SPOCO@0.8 & 0.057  \\
\hline
\end{tabularx}
\end{center}
\caption{Evaluation on a 3D LM dataset of Arabidopsis Ovules \cite{plantseg2020}.  The Adapted Rand Error (ARand error) averaged over the 7 test set 3D stacks is reported. Bottom part of the table shows the scores achieved in weakly supervised settings.}
\label{tab:ovules}
\end{table}

Table~\ref{tab:ovules_Stem} shows SPOCO performance in a transfer learning setting, when fine-tuning a network trained on the Ovules dataset to segment the Stem dataset. The Ovules network trained only on source data does not perform very well, but just 5\% of the target groundtruth annotations brings a two-fold improvement in segmentation accuracy. 
Results in tables~\ref{tab:ovules}~and~\ref{tab:ovules_Stem} are based on HDBSCAN ($min\_size=550$) clustering.

Qualitative results are shown in Fig.~\ref{fig:light_microscopy} (bottom row). Note how the output embeddings from the Ovules network fine-tuned with just 1\% of cells from the target dataset are less crisp due to the domain gap, but the clustering is still able to segment them correctly.

\begin{table}
\begin{center}
\begin{tabularx}{0.8\columnwidth}{|>{\hsize=1.3\hsize\linewidth=\hsize}X|>{\hsize=0.7\hsize\linewidth=\hsize}X|}
\hline
Method & ARand error\\
\hline
Stem only   & 0.074 \\
Ovules only & 0.227 \\
\hline
Ovules+Stem@0.01 & 0.141 $\pm$ 0.002 \\
Ovules+Stem@0.05 & 0.109 $\pm$ 0.002 \\
Ovules+Stem@0.1 & 0.106 $\pm$ 0.004 \\
Ovules+Stem@0.4 & 0.093 $\pm$ 0.003 \\
\hline
\end{tabularx}
\end{center}
\caption{Evaluation on a 3D LM dataset in a transfer learning setting. Ovules dataset acts as the source domain, Stem dataset as target. Performance is shown as Adapted Rand Error, lower is better.
Mean $\pm$ SD are reported across 3 random samplings of the instances from the target dataset. }
\vspace{-1em}
\label{tab:ovules_Stem}
\end{table}

\begin{figure}[!htbp]
\vspace{1.5em}
\begin{center}
\includegraphics[width=1.0\linewidth]{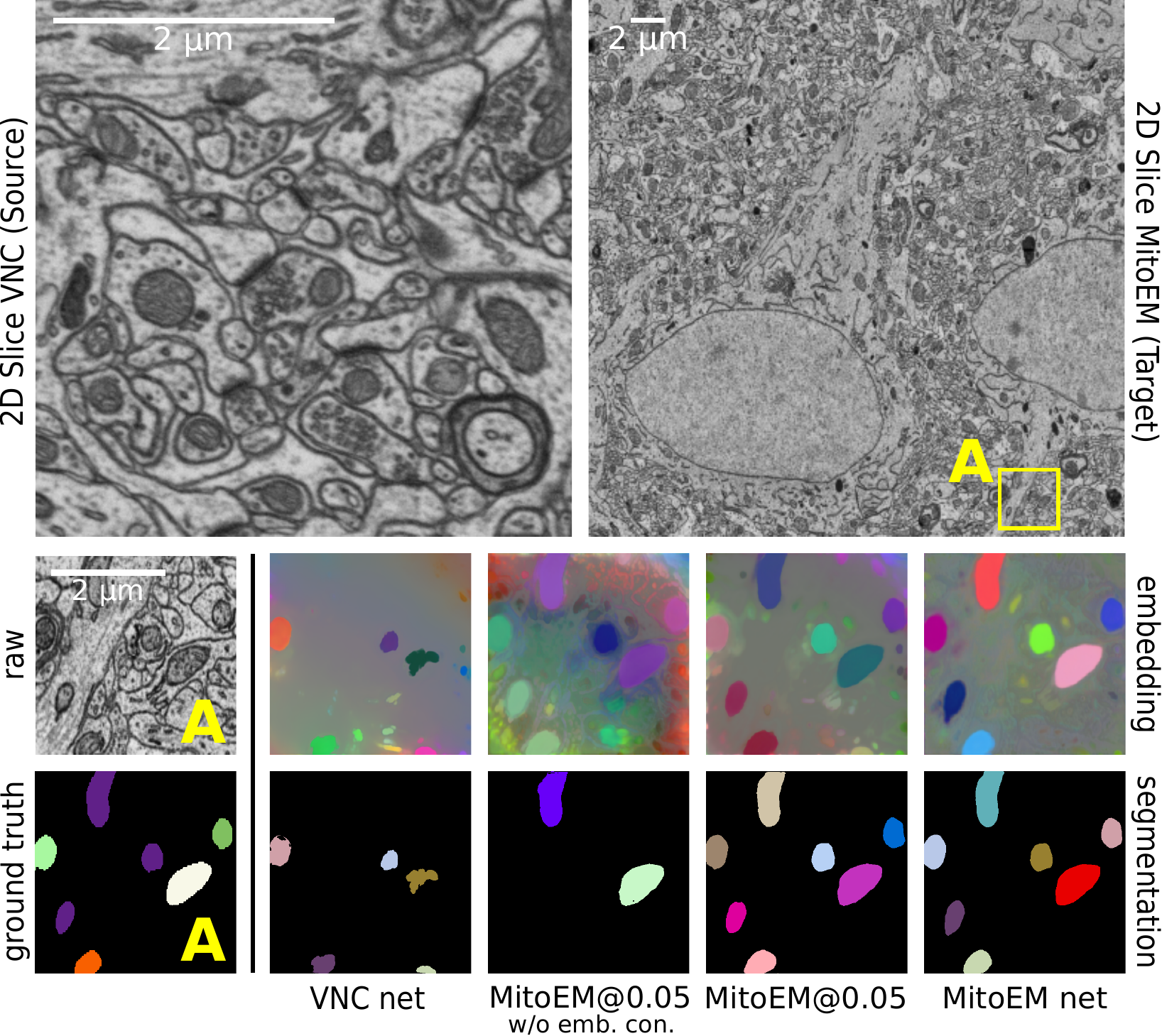}
\end{center}
\vspace{-0.5em}
   \caption{EM segmentation in transfer learning setting. \textbf{\textsc{Top)}} samples from the source (VNC, left) and target (MitoEM, right) datasets; \textbf{\textsc{Middle)}} the input image and the RGB-projected embeddings: trained on VNC only, VNC-pretrained + MitoEM@0.05-finetuned without the embedding consistency, same but with the embedding consistency, trained on MitoEM only; \textbf{\textsc{Bottom)}} groundtruth and predicted segmentations.}
\vspace{-0.8em}
\label{fig:mitoem}
\end{figure}

\textbf{EM datasets.} 
Table~\ref{tab:vnc_mitoem} continues the evaluation of SPOCO performance in a transfer learning setting. We report the average precision at 0.5 IoU threshold (AP@0.5) and the mean average precision (mAP). Similar to the LM case, just  1\% of annotated  objects in the target dataset bring a 1.5 fold improvement in the mean average precision compared to the network trained on source VNC domain only.
A comparison to a network trained only on MitoEM (Table~\ref{tab:vnc_mitoem} bottom) shows that fine-tuning does significantly improve performance for low amounts of training data (1\% of the target). With 10\% of the annotated objects, fine-tuned VNC network does not reach the performance of the SPOCO@0.1 trained directly on MitoEM (target) due to reduced learning rate.
Figure~\ref{fig:mitoem} illustrates the EM experiments. The VNC-net only partially recovers 4 out of 7 groundtruth instances and also produces a false positive. MitoEM@0.05 without consistency loss only recovers 2 instances, while the version with the consistency loss recovers the correct segmentation. Embeddings clustered with HDBSCAN ($min\_size=600$).

\begin{table}[!htbp]
\begin{center}
\begin{tabular}{|l|l|l|}
\hline
Method & AP@0.5 & mAP \\
\hline
VNC     & 0.234 & 0.137  \\
\hline
VNC-MitoEM & & \\
\hdashline
SPOCO@0.01 & 0.368 $\pm$ 0.022 & 0.247 $\pm$ 0.022 \\
SPOCO@0.05 & 0.398 $\pm$ 0.007 & 0.277 $\pm$ 0.006 \\
SPOCO@0.10 & 0.389 $\pm$ 0.013 & 0.268 $\pm$ 0.007 \\
\hline
MitoEM & & \\
\hdashline
SPOCO@0.01 & 0.088 $\pm$ 0.045 & 0.046 $\pm$ 0.025 \\
SPOCO@0.05 & 0.403 $\pm$ 0.055 & 0.280 $\pm$ 0.046 \\
SPOCO@0.10 & 0.481 $\pm$ 0.008 & 0.340 $\pm$ 0.007 \\
SPOCO & 0.560 & 0.429  \\
\hline
\end{tabular}
\end{center}
\caption{Evaluation on MitoEM dataset (target) fine-tuned from the VNC net (upper part) and trained from scratch (lower part). The  performance is measured through average precision (AP@0.5, mAP). Mean $\pm$ SD are reported across 3 random samplings of the instances from the target dataset.}
\vspace{-0.8em}
\label{tab:vnc_mitoem}
\end{table}

\section{Conclusion}
We presented a novel approach to weak supervision for instance segmentation tasks which enables training in a positive unlabeled setting. Here, only a subset of object masks are annotated with no annotations in the background and the loss is applied directly to the annotated objects via a differentiable instance selection step. The unlabeled areas of the images contribute to the training through an instance-level consistency loss. 

We demonstrate the advantage of single-instance losses in a fully supervised setting, reaching state-of-the-art performance on the CVPPP benchmark and improving on strong baselines in several microscopy datasets. Weak positive unlabeled supervision is evaluated on the Cityscapes instance segmentation task and on biological datasets from light and electron microscopy, 2D and 3D, in direct training and in transfer learning. In all cases, the network demonstrates strong segmentation performance at a very reduced manual annotation cost.

In the future, we plan to explore the possibility of fully self-supervised pre-training using the consistency loss and an extended set of augmentations. This would open up the possibility for efficient fine-tuning of the learned feature space with point supervision for both semantic and instance segmentation tasks.

\textbf{Limitations.}
The main drawback of the proposed approach is the lack of a universal clustering method to assign instance labels to pixels based on their embeddings. The existing methods all have benefits and drawbacks; there is no consistent winner that would work robustly across all segmentation benchmarks. Appendix A.4 contains a detailed comparison of different clustering algorithms.

\appendix
\section{Appendix}

\subsection{Network architectures and training parameters}
\begin{figure}[H]
\begin{center}
\includegraphics{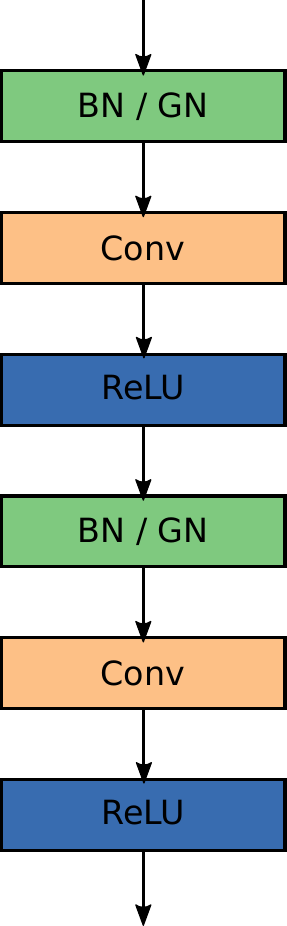}

\end{center}
   \caption{ConvBlock architecture. We use \textit{ConvBlock}, \textit{ConvBlockBN} or \textit{ConvBlockGN} to refer to the block where \textit{no}, \textit{batch} or \textit{group} normalization is used respectively.}
\label{fig:conv_block} 
\end{figure}

The structure of the U-Net embedding network used for each dataset is described using the buliding block shown in Figure~\ref{fig:conv_block}. The number of ConvBlocks in the encoder/decoder part of U-Net is chosen such that the receptive field of features in the last encoder layer is equal to or slightly bigger than the input size. We used group normalization \cite{groupnorm2018} for 3D and electron microscopy experiments and batch normalization for CVPPP and Cityscapes datasets \cite{batchnorm2015}. 

In the tables describing the architecture: second number after the comma corresponds the number of output channels from each layer, \textit{Upsample} denotes nearest-neighbor upsampling and \textit{Concat} denotes channel-wise concatenation of the output for a given decoder layer with the output from the  corresponding encoder layer.

Unless otherwise specified, Adam optimizer \cite{kingma2014adam} with an initial learning rate of $0.0002$, weight decay $10^{-5}$, $\beta_1=0.9$ and $\beta_2=0.999$ was used for training. Learning rate was reduced by a factor of $0.2$ when the validation loss stopped improving after a dataset-dependent number of iterations. Training was stopped when the learning rate dropped below $10^{-6}$ or maximum number of iterations was reached.
In all our experiments we use 16-dimensional embedding space, i.e. the output from the U-Net after the last 1$\times$1 convolution has 16 channels. Input images were globally normalized to zero mean and a standard deviation of one unless stated otherwise.

\paragraph{CVPPP.}
Table~\ref{tab:cvppp_arch} shows the 2D U-Net architecture used in the experiment. All networks were trained for up to 80K iterations (unless the stopping criteria was not satisfied before) with a minibatch size of 4. Input images were randomly scaled, flipped horizontally and vertically and cropped to 448$\times$448 pixels. Before passing to $f(\cdot)$ and $g(\cdot)$ networks, random color jitter and Gaussian blur were applied.

\begin{table}[H]
\begin{center}
\begin{tabular}{c}
\hline\hline
3-channel image $x \in \mathbb{R}^{M \times N \times 3}$ \\
\hline
ConvBlockBN, 16, MaxPool 2$\times$2 \\
\hline
ConvBlockBN, 32, MaxPool 2$\times$2 \\
\hline
ConvBlockBN, 64, MaxPool 2$\times$2 \\
\hline
ConvBlockBN, 128, MaxPool 2$\times$2  \\
\hline
ConvBlockBN, 256, MaxPool 2$\times$2  \\
\hline
ConvBlockBN, 512, Upsample 2$\times$2 \\
\hline
Concat, 256 + 512 \\
\hline
ConvBlockBN, 256, Upsample 2$\times$2 \\
\hline
Concat, 128 + 256 \\
\hline
ConvBlockBN, 128, Upsample 2$\times$2 \\
\hline
Concat, 64 + 128 \\
\hline
ConvBlockBN, 64, Upsample 2$\times$2 \\
\hline
Concat, 32 + 64 \\
\hline
ConvBlockBN, 32, Upsample 2$\times$2 \\
\hline
Concat, 16 + 32 \\
\hline
ConvBlockBN, 16, conv 1$\times$1, $d$ \\
\hline\hline
\end{tabular}
\end{center}
\caption{U-Net architecture for CVPPP and Cityscapes datasets. $(M,N)=(448,448)$, $d=16$ for CVPPP and $(M,N)=(384, 768)$, $d=8$ for Cityscapes.}
\label{tab:cvppp_arch}
\end{table}

\paragraph{Cityscapes.}
See Table~\ref{tab:cvppp_arch} for an overview of 2D U-Net architecture for the Cityscapes semantic instance segmentation task. All networks were trained for up to 90K iterations with a minibatch size of 16. The network output dimension was set to 8. Input images were randomly cropped to 358$\times$768 patches. Random flipping and scaling (ratio in [0.5, 2.0]), Gaussian blurring, color jitter and random conversion to grayscale was applied to the input before passing it to $f(\cdot)$ and $g(\cdot)$ networks. 

\paragraph{Light microscopy datasets.}
3D U-Net architecture used for the light microscopy datasets is shown in Table~\ref{tab:ovules_arch}. Ovules networks were trained for up to 200K iterations (or until the stopping criteria was satisfied) with a minibatch size of 8. Stem networks were fine-tuned with a fixed, reduced learning rate of $0.00002$ for 100K iterations. 3D patches of shape 40$\times$64$\times$64 (\textit{ZYX} axes ordering) were used. Patches were augmented with random rotations, flips and elastic deformations. Gaussian noise was added to the input before passing through $f(\cdot)$ and $g(\cdot)$ networks. \\
For a  fair comparison with other methods we do not stitch the patches to recover the whole volume, but evaluate on the patch-by-patch basis.

\begin{table}
\begin{center}
\begin{tabular}{c}
\hline\hline
1-channel 3D patch $x \in \mathbb{R}^{K \times M \times M \times 1}$ \\
\hline
ConvBlockGN, 64, MaxPool 2$\times$2 \\
\hline
ConvBlockGN, 128, MaxPool 2$\times$2  \\
\hline
ConvBlockGN, 256, MaxPool 2$\times$2  \\
\hline
ConvBlockGN, 512, Upsample 2$\times$2 \\
\hline
Concat, 256 + 512 \\
\hline
ConvBlockGN, 256, Upsample 2$\times$2 \\
\hline
Concat, 128 + 256 \\
\hline
ConvBlockGN, 128, Upsample 2$\times$2 \\
\hline
Concat, 64 + 128 \\
\hline
ConvBlockGN, 64, conv 1$\times$1, $d=16$ \\
\hline\hline
\end{tabular}
\end{center}
\caption{U-Net architecture for Ovules and Stem datasets, $K = 40, M=64$. All convolutions and max pooling operations are 3D.}
\label{tab:ovules_arch}
\end{table}

\begin{table}
\begin{center}
\begin{tabular}{c}
\hline\hline
1-channel image $x \in \mathbb{R}^{M \times M \times 1}$ \\
\hline
ConvBlockGN, 16, MaxPool 2$\times$2 \\
\hline
ConvBlockGN, 32, MaxPool 2$\times$2 \\
\hline
ConvBlockGN, 64, MaxPool 2$\times$2 \\
\hline
ConvBlockGN, 128, MaxPool 2$\times$2  \\
\hline
ConvBlockGN, 256, MaxPool 2$\times$2  \\
\hline
ConvBlockGN, 512, MaxPool 2$\times$2 \\
\hline
ConvBlockGN, 1024, Upsample 2$\times$2 \\
\hline
Concat, 512 + 1024 \\
\hline
ConvBlockGN, 512, Upsample 2$\times$2 \\
\hline
Concat, 256 + 512 \\
\hline
ConvBlockGN, 256, Upsample 2$\times$2 \\
\hline
Concat, 128 + 256 \\
\hline
ConvBlockGN, 128, Upsample 2$\times$2 \\
\hline
Concat, 64 + 128 \\
\hline
ConvBlockGN, 64, Upsample 2$\times$2 \\
\hline
Concat, 32 + 64 \\
\hline
ConvBlockGN, 32, Upsample 2$\times$2 \\
\hline
Concat, 16 + 32 \\
\hline
ConvBlockGN, 16, conv 1$\times$1, $d=16$ \\
\hline\hline
\end{tabular}
\end{center}
\caption{U-Net architecture for VNC and MitoEM datasets, $M=448$.}
\label{tab:mitos_arch}
\end{table}

\paragraph{Electron microscopy datasets.}
2D U-Net architecture for the VNC and MitoEM datasets is shown in Table~\ref{tab:mitos_arch}. The source VNC network was trained for up to 100K iterations with a minibatch size of 4. MitoEM networks were fine-tuned with a fixed, reduced learning rate of $0.00002$ for 100K iterations. 2D patches of shape 448$\times$448 were used. Patches were augmented with random rotations, flips and elastic deformations. Gaussian noise was added to the input before passing through $f(\cdot)$ and $g(\cdot)$ networks.

\paragraph{Adversarial training.}
As mentioned in Sec.\ref{ssec:single_obj} our approach can be used for adversarial training, where the pixel embedding network is a generator of object masks, which the discriminator learns to distinguish from the ground truth object segmentation masks. In this case the embedding network (generator) is trained with the following objective:
\begin{equation}
\label{eq:wgan}
L_{adv} = L_{SO} + \lambda L_{wgan}
\end{equation}
with $L_{SO}$ defined in Eq.~\ref{eq:cl_dice}. 
For adversarial training we use Wasserstein GAN with gradient penalty (WGAN-GP) \cite{NIPS2017_wgan-gp} objective function given by:

\begin{equation}
\label{eq:wgan_d}
\begin{aligned}
V_D(G, D) & = \mathbb{E}_{\mathbf{\tilde{x}} \sim \mathbb{P}_g} [D(\mathbf{\tilde{x}})] - \mathbb{E}_{\mathbf{x} \sim \mathbb{P}_r} [D(\mathbf{x})] \\
& + \lambda \mathbb{E}_{\mathbf{\hat{x}} \sim \mathbb{P}_{\mathbf{\hat{x}}}} [(\lVert \nabla_{\mathbf{\hat{x}}} D(\mathbf{\hat{x}}) \lVert - 1)^2]
\end{aligned}
\end{equation}

\begin{equation}
\label{eq:wgan_g}
L_{wgan} = V_G(G, D) = - \mathbb{E}_{\mathbf{\tilde{x}} \sim \mathbb{P}_g} [D(\mathbf{\tilde{x}})]
\end{equation}
for the critic and the embedding network (generator) respectively. $\mathbb{P}_r$ is the distribution of ground truth mask, $\mathbb{P}_g$ is the distribution of predicted "soft" masks and $\mathbb{P}_{\mathbf{\hat{x}}}$ is the sampling distribution.
Table~\ref{tab:wgan_critic} shows the architecture of the critic used in the Ovules dataset experiments. 
Our final objective for the embedding netowrk is given by: $L_{SO} + \zeta L_{wgan}$ or $L_{SSO} + \zeta L_{wgan}$ depending on whether full or sparse supervision is used.
Training of the embedding network and the critic is done using the Adam optimizer with $\beta_1=0.5$, $\beta_2=0.9$ and initial learning rate of $0.0001$ for both networks.
We use $n_{critic}=5$ iterations per each iteration of the embedding network. We use $\lambda=10$ (GP weight) and $\zeta=0.1$ ($L_{wgan}$ weight) in our experiments. In order to prevent uninformative gradients from the critic at the beginning of the training process, $L_{wgan}$ is enabled after the warm-up period of 50K iterations.

In our experiments, adversarial training does not by itself bring a significant performance improvement over the Dice-based loss. Both losses can also be used in combination which can be beneficial: as we show in Table~\ref{tab:ovules}, the combined loss outperforms a much more complex 3-step state-of-the-art segmentation pipeline. 
This finding is similar to \cite{luc2016semantic} where authors use an adversarial approach to train a semantic segmentation model. Our approach differs, as in our setup the discriminator focuses more on the individual object properties instead of the global statistics of the semantic mask predicted by the network.

\begin{table}[H]
\begin{center}
\begin{tabular}{c}
\hline\hline
1-channel image $x \in [0, 1]^{K \times M \times M \times 1}$ \\
\hline
ConvBlock, 64, MaxPool 2$\times$2 \\
\hline
ConvBlock, 128, MaxPool 2$\times$2  \\
\hline
ConvBlock, 256, MaxPool 2$\times$2  \\
\hline
ConvBlock, 512, Upsample 2$\times$2 \\
\hline
dense layer, 1 \\
\hline\hline
\end{tabular}
\end{center}
\caption{WGAN-GP critic architecture used in the adversarial setting on Ovules dataset, $K = 40, M=64$. All convolutions and max pooling operations are 3D. No batch or group normalization was used. ReLU was replaced by leaky ReLU activation function with $\alpha = 10^{-2}$.}
\label{tab:wgan_critic}
\end{table}

\subsection{Ablation study of loss functions}

In this section, we study the impact of different loss variants and the choice of $g(\cdot)$ network on the final performance of our method.
Table~\ref{tab:loss_ablation_cvppp} presents segmentation and counting scores (CVPPP test set) with different loss variants. HDBSCAN with $min\_size=200$ and no foreground mask was used for clustering the network outputs in all cases.
We see that when only a few ground truth objects are used for training (10\%, 40\%) the consistency term $L_{U\_{con}}$ (Eq.~\ref{eq:ul_con}) has a much stronger impact on the final segmentation performance than the unlabeled "push" term $L_{U\_push}$ (Eq.~\ref{eq:ul_push}).  The absence of  $L_{U\_push}$ term worsens the segmentation and counting scores in all experiments.

\begin{table}[!htbp]
\begin{center}
\begin{tabularx}{\columnwidth}{|>{\hsize=1.0\hsize\linewidth=\hsize}X|>{\hsize=0.6\hsize\linewidth=\hsize}X|>{\hsize=0.4\hsize\linewidth=\hsize}X|}
\hline
Loss function & SBD & $|DiC|$ \\
\hline
@0.1 & 0.788 $\pm$ 0.017 & 5.4 $\pm$ 0.3 \\
@0.1 w/o $L_{U\_push}$ & 0.734 $\pm$ 0.042 & 8.5 $\pm$ 0.4 \\
@0.1 w/o $L_{U\_con}$ & 0.720 $\pm$ 0.037 & 6.3 $\pm$ 0.1 \\
\hline
@0.4 & 0.824 $\pm$ 0.003 & 3.2 $\pm$ 0.5 \\
@0.4 w/o $L_{U\_push}$ & 0.779 $\pm$ 0.045 & 3.0 $\pm$ 0.7 \\
@0.4 w/o $L_{U\_con}$ & 0.738 $\pm$ 0.019 & 2.1 $\pm$ 0.1 \\
\hline
@0.8 & 0.828 $\pm$ 0.010 & 1.6 $\pm$ 0.2 \\
@0.8 w/o $L_{U\_push}$ & 0.797 $\pm$ 0.014 & 1.9 $\pm$ 0.4 \\
@0.8 w/o $L_{U\_con}$ & 0.810 $\pm$ 0.010 & 2.1 $\pm$ 0.2 \\
\hline
\end{tabularx}
\end{center}
\caption{Ablation study of different loss variants. We report segmentation (SBD) and counting ($|DiC|$) scores on the CVPPP test set. Ablation of the $L_{U\_con}$ $L_{U\_push}$ term in the semi-supervised setting is reported for 10\%, 40\% and 80\% of randomly selected ground truth objects. Mean $\pm$ SD are reported across 3 random samplings of the ground truth objects.}
\label{tab:loss_ablation_cvppp}
\end{table}

To finalize the ablation study, we trained the network in a fully-supervised setting using the consistency regularization from the weakly-supervised setup.
Tab~\ref{tab:full_supp_abl} shows the comparison between all four variants.
Using the consistency term $L_{U\_con}$ together with the instance-based term $L_{obj}$ on the Cityscapes validation set gives the highest mAP@0.5 score of 0.459 as compared to 0.387 (discriminative loss \cite{debrabandere2017semantic}), 0.418 (\cite{debrabandere2017semantic} + $L_{obj}$) and 0.429 (\cite{debrabandere2017semantic} + $L_{U\_con}$).
For CVPPP the SBD metric on the validation set improves from 0.847 (full supervision as in \cite{debrabandere2017semantic}) to 0.852 (\cite{debrabandere2017semantic} + $L_{obj}$), to 0.853 (\cite{debrabandere2017semantic} + $L_{obj}$) and to 0.849 (\cite{debrabandere2017semantic} + $L_{obj}$ + $L_{U\_con}$).

\begin{table}[!htbp]
\begin{center}
\begin{tabular}{|l|l|l|}
\hline
Loss function & CVPPP & Cityscapes \\
\hline
DL \cite{debrabandere2017semantic}  & 0.847 & 0.387 \\
DL + $L_{obj}$                      & 0.852 & 0.418 \\
DL + $L_{U\_con}$                   & \textbf{0.853} & 0.429 \\
DL + $L_{obj}$ + $L_{U\_con}$       & 0.849 & \textbf{0.459} \\
\hline
\end{tabular}
\end{center}
\vspace{-0.8em}
\caption{
Weakly-supervised regularization in the fully-supervised setting. SBD (CVPPP datasets) and mAP@0.5 (Cityscapes dataset) computed on the validation sets.}
\label{tab:full_supp_abl}
\end{table}

\subsection{g-network ablations}
\label{ssec:g_net_ablation}
We experiment with different types of the $g$ network used in the consistency loss to better understand its effect. Apart from the momentum $g$ described in Sec.~\ref{ssec:sparse_single_obj} we consider three other variants: (1) weights are \textit{shared} between $f$ and $g$, i.e. $\theta_g = \theta_f$, (2) $g$ shares the weights with $f$, but uses spatial \textit{dropout} \cite{Tompson_2015_CVPR} in the bottleneck layer of the U-Net architecture, (3) $g$ uses independent set of weights \textit{trained} by back-propagation. \\
For the purpose of this ablation, we split the CVPPP A1 training set into 103 randomly selected images used for training and report the results on the remaining 25 images.

\begin{figure}[!htbp]
\begin{center}
\includegraphics[width=1.0\linewidth]{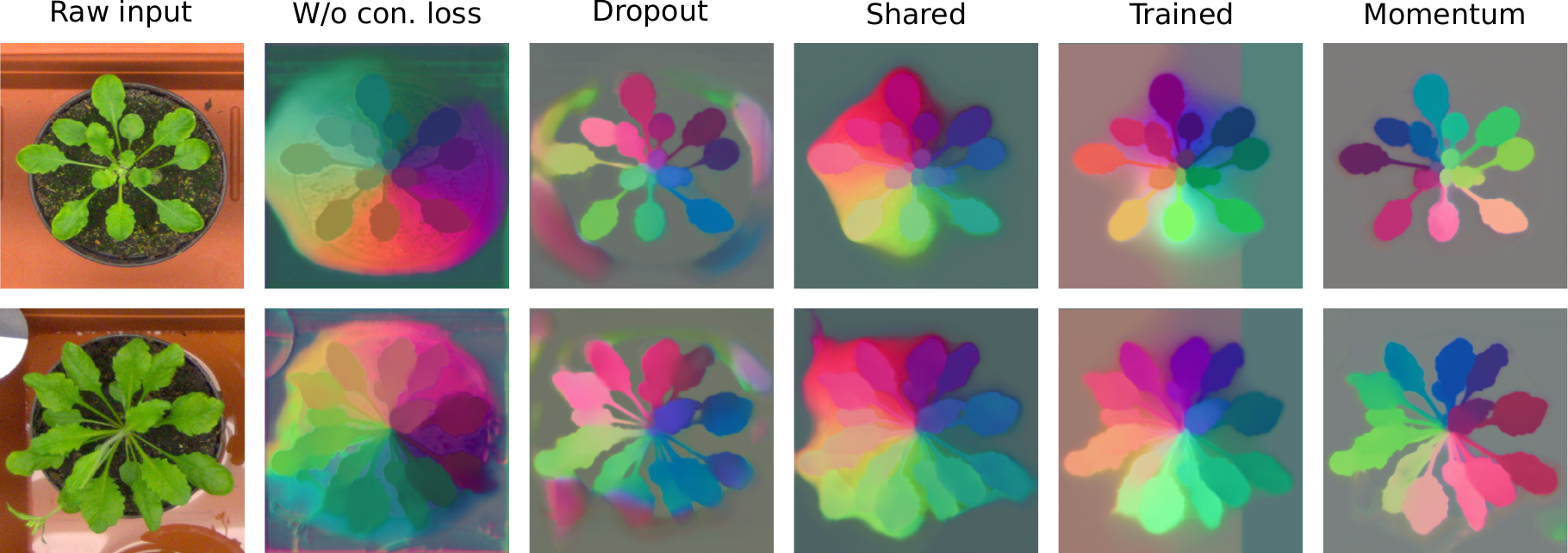}
\end{center}
\vspace{-0.5em}
   \caption{Qualitative comparison of the PCA-projected outputs from the network $f$ for different training setups: (col 2) no consistency term, (col 3) dropout $g$, (col 4) shared $g$, (col 5) trained $g$, (col 6) momentum $g$. Two images from the CVPPP test set were randomly selected. SPOCO@0.1 was used for training.}
\label{fig:g_net_qualitative}
\end{figure}

Table~\ref{tab:g_net_ablation_cvppp} shows the segmentation and counting scores for the HDBSCAN-clustered embeddings together with training dynamics for different variants of $g$. "Dropout" variant shows a good initial convergence rate, but overfits quickly and has the worse final performance. "Trained" and "shared" variants show comparable average scores, however the variance is much larger for the "trained", which is prone to training instabilities. "Momentum" outperforms the others by a large margin and has the fastest convergence speed.
Figure~\ref{fig:g_net_qualitative} shows the PCA-projected netork outputs for two randomly selected CVPPP images (test set) and five different settings.
One can see that sparse training without the consistency loss (column 2) fails to separate the background. Using dropout $g$ leads to artifacts in the unlabeled region. "Shared" and "trained" variants of $g$ (columns 4 and 5) provide limited background separation, but fail to produce crisp embeddings. The "momentum" variant (column 6) is able to correctly separate the background.

\begin{table}[!htbp]
\begin{center}
\begin{tabularx}{\columnwidth}{|>{\hsize=1.0\hsize\linewidth=\hsize}X|>{\hsize=0.6\hsize\linewidth=\hsize}X|>{\hsize=0.4\hsize\linewidth=\hsize}X|}
\hline
g-network & SBD & $|DiC|$ \\
\hline
Shared & 0.602 $\pm$ 0.016 & 6.0 $\pm$ 0.6 \\
Dropout & 0.507 $\pm$ 0.060 & 7.9 $\pm$ 0.9 \\
Trained & 0.591 $\pm$ 0.131 & 5.7 $\pm$ 2.0 \\
Momentum & \textbf{0.649 $\pm$ 0.045 } & \textbf{4.5 $\pm$ 1.6} \\
\hline
\end{tabularx}
\vspace{.5em}
\begin{flushleft}
\vspace{-1em}
\includegraphics[width=.95\linewidth]{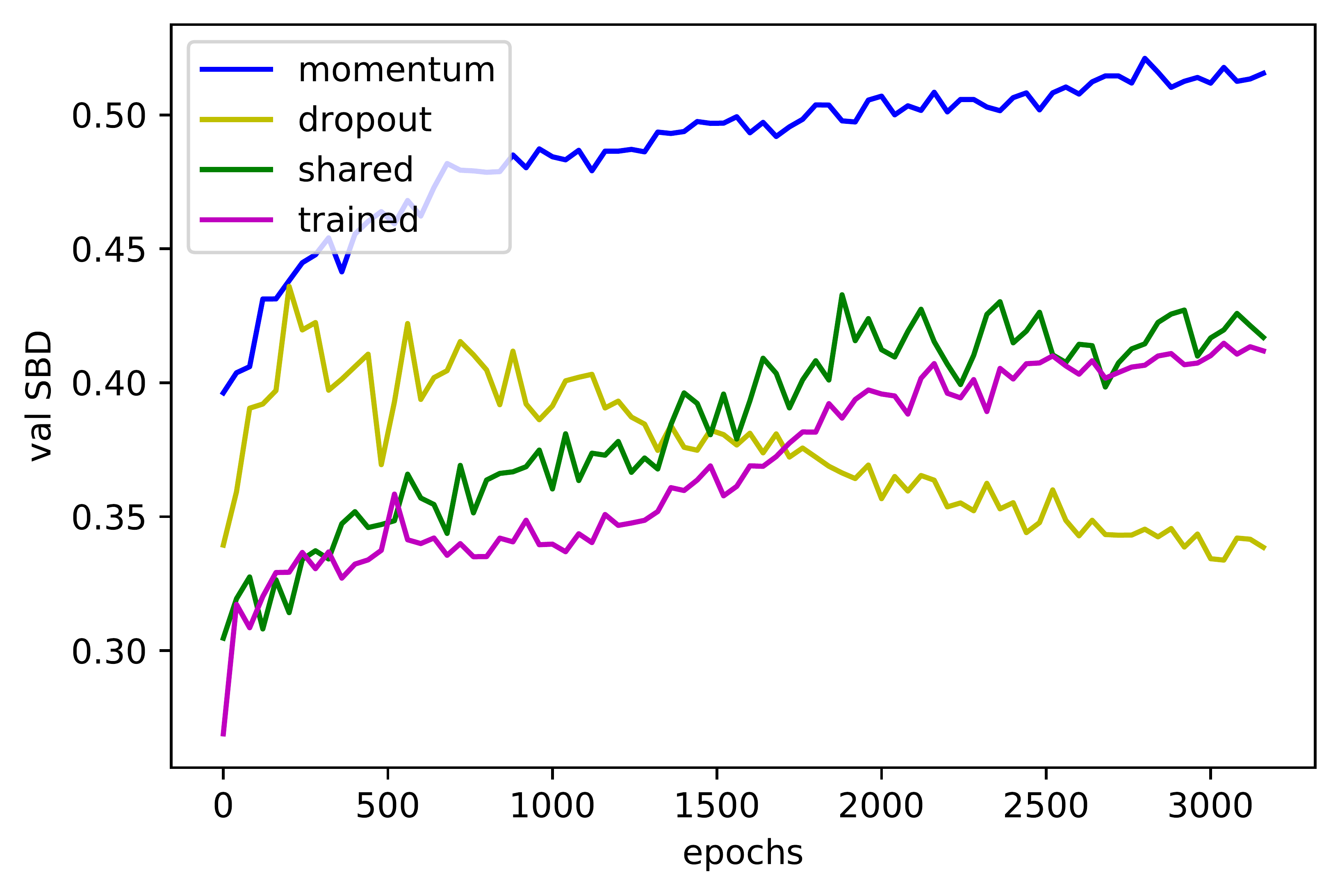}	
\end{flushleft}
\vspace{-1em}
\end{center}
\caption{(Top) Segmentation performance computed for different types of g-network on the CVPPP validation set. (Bottom) Comparison between g-network variants during training. SPOCO@0.1 used for training, mean $\pm$ SD across 3 random samplings of the groundtruth objects is shown.}
\label{tab:g_net_ablation_cvppp}
\vspace{-1.5em}
\end{table}

In transfer learning setting, the embedding network trained on the source domain 
 is fine-tuned on the target domain with just a few groundtruth objects.
Results on the EM data, where VNC dataset \cite{zora91121} is the source and the MitoEM \cite{wei2020mitoem} is the target domain (Tab~\ref{tab:vnc_mitoem_ablation}) show significant drop in segmentation scores across all experiments if the embedding consistency is removed from the loss.

\begin{table}[!htbp]
\begin{center}
\begin{tabular}{|l|l|l|}
\hline
Method & AP@0.5 & mAP \\
\hline
@0.01 & 0.368 $\pm$ 0.022 & 0.247 $\pm$ 0.022 \\
@0.01 w/o $L_{U\_con}$ & 0.306 $\pm$ 0.014 & 0.210 $\pm$ 0.008 \\
\hdashline
@0.05 & 0.398 $\pm$ 0.007 & 0.277 $\pm$ 0.006 \\
@0.05 w/o $L_{U\_con}$ & 0.319 $\pm$ 0.002 & 0.227 $\pm$ 0.002 \\
\hdashline
@0.10 & 0.389 $\pm$ 0.013 & 0.268 $\pm$ 0.007 \\
@0.10 w/o $L_{U\_con}$ & 0.301 $\pm$ 0.012 & 0.212 $\pm$ 0.007 \\
\hline
\end{tabular}
\end{center}
\caption{Ablation of the consistency term $L_{U\_con}$ in the transfer learning setting with 1\%, 5\%, 10\% of groundtruth objects (target domain). Average precision measured on the target task of MitoEM mitochondria segmentation is reported. VNC dataset serves as a source domain. Mean $\pm$ SD are reported across 3 random samplings of the instances from the target dataset.}
\label{tab:vnc_mitoem_ablation}
\end{table}

\subsection{Comparison of clustering algorithms}
Apart from the standard mean-shift and HDBSCAN, we introduce two additional clustering methods: (1) a hybrid scheme called \textit{consistency clustering} and (2) a fast affinity graph partitioning. Consistency clustering (Algorithm~\ref{alg:clustering}) works by passing two augmented versions of the input through the networks $f$ and $g$, producing embeddings $\mathcal{E}_f$ and $\mathcal{E}_g$ respectively. We cluster $\mathcal{E}_f$ using mean-shift with bandwidth set to the pull force margin $\delta_v$. Then for each segmented object $S_k$, we randomly select $M$ anchor points and for each anchor we extract a new object $\hat{S}_k^m$ by taking a $\delta_v$-neighborhood around the anchor in the $\mathcal{E}_g$ space. If the median intersection-over-union (IoU) between $S_k$ and each of the $\hat{S}_k^m$ objects is lower than a predefined threshold, we discard $S_k$ from the final segmentation. This is based on the premise that clusters corresponding to the real objects should remain consistent between $\mathcal{E}_f$ and $\mathcal{E}_g$.

\begin{algorithm}[ht]
\KwIn{Set of mean-shift segmented objects $\mathcal{S}$, embeddings from the  g-network $\mathcal{E}_g = \{\boldsymbol{e}_0, \boldsymbol{e}_1, ..., \boldsymbol{e}_N\}$, IoU threshold $t_{IoU}$, number of anchors per object to sample $M$}
\KwOut{New set of segmented objects $\hat{\mathcal{S}}$}
$\hat{\mathcal{S}} = \{\}$\;
\For{$S_k \in \mathcal{S}$}{
  $\mathcal{A}_k = \{\boldsymbol{a}_k^1,..., \boldsymbol{a}_k^M \mid \boldsymbol{a}_k^m \in \mathcal{E}_g\} \textrm{ - anchors of} ~ S_k$\;
  $I_{IoU} = \{ \}$\;
  \For{$\boldsymbol{a}_k^m \in \mathcal{A}_k$}{
    $\hat{S}_k^m = \{s_i \mid s_i= \|\boldsymbol{e}_i - \boldsymbol{a}_k^m \| < \delta_v \}$\;
    $I_{IoU} \cup \textrm{IoU}(\hat{S}_k^m, S_k)$\;
  }
  \If{$\textrm{med}(I_{IoU}) > t_{IoU}$} {
    $\hat{\mathcal{S}} = \hat{\mathcal{S}} \cup \{ S_k \} $;
  }
}
\Return $\hat{S}$;
\caption{Consistency clustering} \label{alg:clustering}
\end{algorithm}

The affinity graph-based method proceeds similar to \cite{Lee2021}: we convert the embedding space into a graph partitioning problem by introducing a grid-graph that contains a node for each pixel and connects all direct neighbor pixel via edges.
Following \cite{lee2017superhuman} and \cite{mws2018} we introduce additional long-range edges that connect pixels that are not direct neighbors in a fixed offset pattern. Following \cite{Lee2021} we derive the edge weight $w_{ij}$, or affinity, between pixel $i$ and $j$ from the embedding vector $\boldsymbol{e}_i$ and $\boldsymbol{e}_j$ via
\begin{equation}
    w_{ij} = 1 - \textrm{max}(\frac{2 \delta_d - \|\boldsymbol{e}_i - \boldsymbol{e}_j\|}{2 \delta_d}, 0)^2.
\end{equation}
Here, $\delta_d$ is the hinge from Eq.~\ref{eq:dist_term} and we use the L2 norm to measure the distance in the embedding space. This weight is derived from the distance term (Eq.~\ref{eq:dist_term}) and is maximally attractive (0) when the embedding distance is zero and becomes maximally repulsive (1) for embedding distances larger than $2 \delta_d$. We obtain an instance segmentation with the Mutex Watershed algorithm \cite{mws2018}, which operates on long-range affinity graphs. We introduce long-range edges between all pixel pairs with distance 3, 9 and 27 across all dimensions. This choice yields good segmentation results empirically; potentially we could obtain even better results with this approach by determining the offset pattern via grid search.

Quantitative comparison of 4 different clustering algorithms: HDBSCAN ($min\_size=200$), Mean-shift (with bandwidth set to $\delta_v=0.5$), Consistency Clustering ($t_{IoU}=0.6$) and affinity-based clustering are shown in Table~\ref{tab:clustering}. We report the segmentation and counting scores as well as runtimes on the CVPPP validation set. We used the embedding networks trained using SPOCO@0.1 (i.e. 10\% of randomly selected ground truth objects) and SPOCO (trained with full supervision).

\begin{figure*}[h]
\begin{center}
\includegraphics[width=1.0\linewidth]{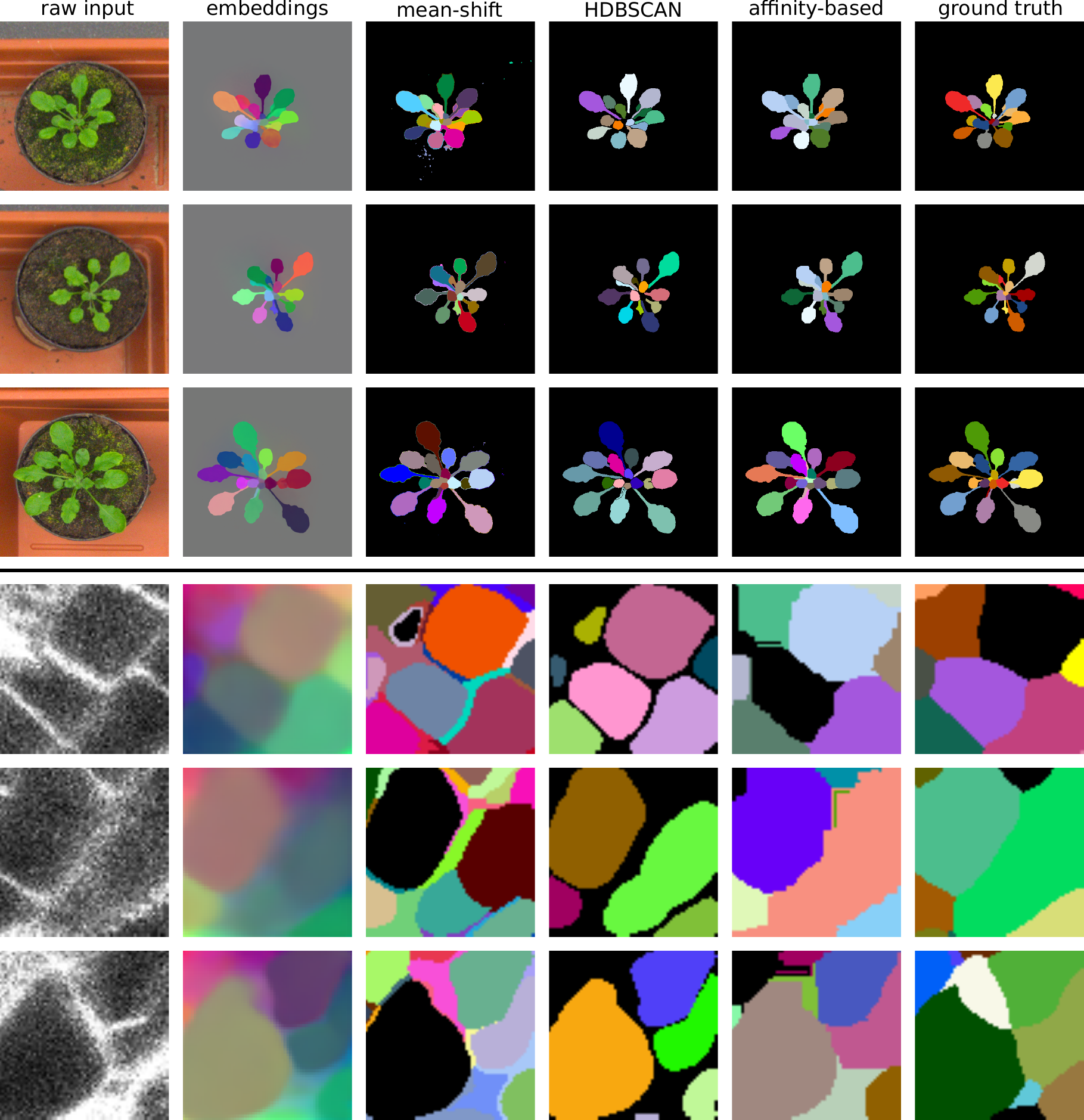}
\end{center}
  \caption{Qualitative comparison of different clustering schemes on the samples from the (top) CVPPP validation set and (bottom) Ovules test set. Fully supervised SPOCO was used to train embeddings.}
\label{fig:clustering}
\end{figure*}

Mean-shift has a high recall (correctly recovers most instances), but low precision (it tends to over-segment the image around the boundary of the objects, see Fig~\ref{fig:clustering}) resulting in high number of false positives and inferior counting scores. Consistency clustering significantly improves the initial mean-shift segmentation resulting in the best segmentation metric for the network trained with weak supervision (SPOCO@0.1).
Affinity-based (Mutex Watershed) and density-based (HDBSCAN) methods have similar segmentation scores, with the former achieving much better counting performance in both full (SPOCO) and weak (SPOCO@0.1) supervision. The affinity-based approach has much lower runtimes compared to the other clustering methods.

\begin{table}[h]
\begin{center}
\begin{tabular}{|l|l|l|l|}
\hline
Method (CVPPP) & SBD & $|DiC|$ & t [s] \\
\hline
SPOCO@0.1 & & & \\
\hdashline
 Consistency    & 0.729 $\pm$ 0.086 &  2.7 $\pm$ 1.7 & 252.3 \\
    HDBSCAN     & 0.653 $\pm$ 0.077 &  5.7 $\pm$ 1.7 & 82.3 \\
 Mean-shift     & 0.356 $\pm$ 0.048 & 20.7 $\pm$ 6.2 & 201.2 \\
 Affinity-based & 0.615 $\pm$ 0.061 &  2.6 $\pm$ 2.3 & 0.45 \\
\hline
SPOCO  & & & \\
\hdashline
   HDBSCAN     & 0.834 &   1.6 & 164.7 \\
Mean-shift     & 0.541 & 10.92 & 121.9 \\
Affinity-based & 0.833 &  0.88 &   0.4 \\
\hline
\end{tabular}
\end{center}

\caption{Performance and runtime comparison of the clustering methods on the CVPPP validation set. We compare the results for SPOCO@0.1, where mean $\pm$ SD are reported across 3 random samplings of the ground truth objects as well as fully supervised SPOCO (bottom), for which we report results from a single training.}
\label{tab:clustering}
\end{table}

\begin{table}[h]
\begin{center}
\begin{tabular}{|l|l|l|}
\hline
Method (Ovules) & Arand error & t [s] \\
\hline
HDBSCAN & 0.133 & 95.036 \\
Mean-shift & 0.102 & 279.202 \\
Affinity-based & 0.086 & 0.955 \\
\hline
\end{tabular}
\end{center}

\caption{Performance (Adapted Rand Error) and runtime comparison of the clustering methods on the ovules test set. Embedding network trained with fully supervised SPOCO.}
\label{tab:clustering2}
\end{table}

Similarly, Table~\ref{tab:clustering2} shows comparison of 3 clustering algorithms: HDBSCAN ($min\_size=600$), Mean-shift ($bandwidth = \delta_v$) and affinity-based on the Ovules test set. We skip the consistency clustering, since the embedding network is trained in the full supervision setting. We notice that for the dense tissue segmentation problems, HDBSCAN classifies the low density areas between the cells as noise, and additional post-processing is required in order to fill the empty space. Results reported in Tab~\ref{tab:ovules} and Fig~\ref{fig:light_microscopy} are based on the watershed post-processing.
Here, for fair comparison with other methods we don't use the watershed post-processing on the HDBSCAN clustering results (see Fig~\ref{fig:light_microscopy} bottom).
Overall, the parameter-free, affinity-based clustering is much faster (3 orders of magnitude faster) than other methods under consideration and provides the best performance-runtime ratio. The downside of HDBSCAN is its sensitivity to the \textit{min\_size} hyperparameter, longer running times and the need for additional post-processing for dense tissue segmentation problems.

Qualitative results on samples from the CVPPP and Ovules datasets are illustrated in Figure~\ref{fig:clustering}.

\subsection{Training with limited annotation budget}
Choosing a fixed annotation budget of $N$ ground truth instances we can objectively compare the weakly supervised training with the dense, fully supervised one. We set $N=16$, which corresponds to roughly 1\% of the objects from the CVPPP training set containing 1683 objects spread across 103 files (we use train/val script described in Sec~\ref{ssec:g_net_ablation}).
In the \textit{dense} setup we randomly choose a single groundtruth file with 16 objects and dense labeling (including the background label), whereas in the \textit{sparse} setting we randomly sample 16 objects from the whole training set, resulting in 16 files, each with only one object labeled. We train fully supervised SPOCO using the densely labeled image and weakly supervised SPOCO using the sparsely labeled images.

Segmentation metrics and embeddings emerging the two training schemes are shown in Tab~\ref{tab:sparse_vs_dense}. The network trained from dense annotations is prone to over-fitting and result in visible artifacts in the embedding space. On the other hand, exposing the network to a much more varied training set in the sparse setting and the presence of a strong consistency regularizer results in a feature space of much better quality. Quantitative comparison confirms that the \textit{sparse} significantly outperforms the \textit{dense} setting in terms of segmentation and counting scores.

\begin{table}[!htbp]
\begin{tabularx}{\columnwidth}{|>{\hsize=1.0\hsize\linewidth=\hsize}X|>{\hsize=0.6\hsize\linewidth=\hsize}X|>{\hsize=0.4\hsize\linewidth=\hsize}X|}
\hline
Training scheme & SBD & $|DiC|$ \\
\hline
1\% dense   & 0.380     & 9.8 \\
1\% sparse  & \textbf{0.691}     & \textbf{2.2} \\
\hline
\end{tabularx}
\vspace{.5em}
\begin{flushleft}
\vspace{-1em}
\includegraphics[width=1.0\linewidth]{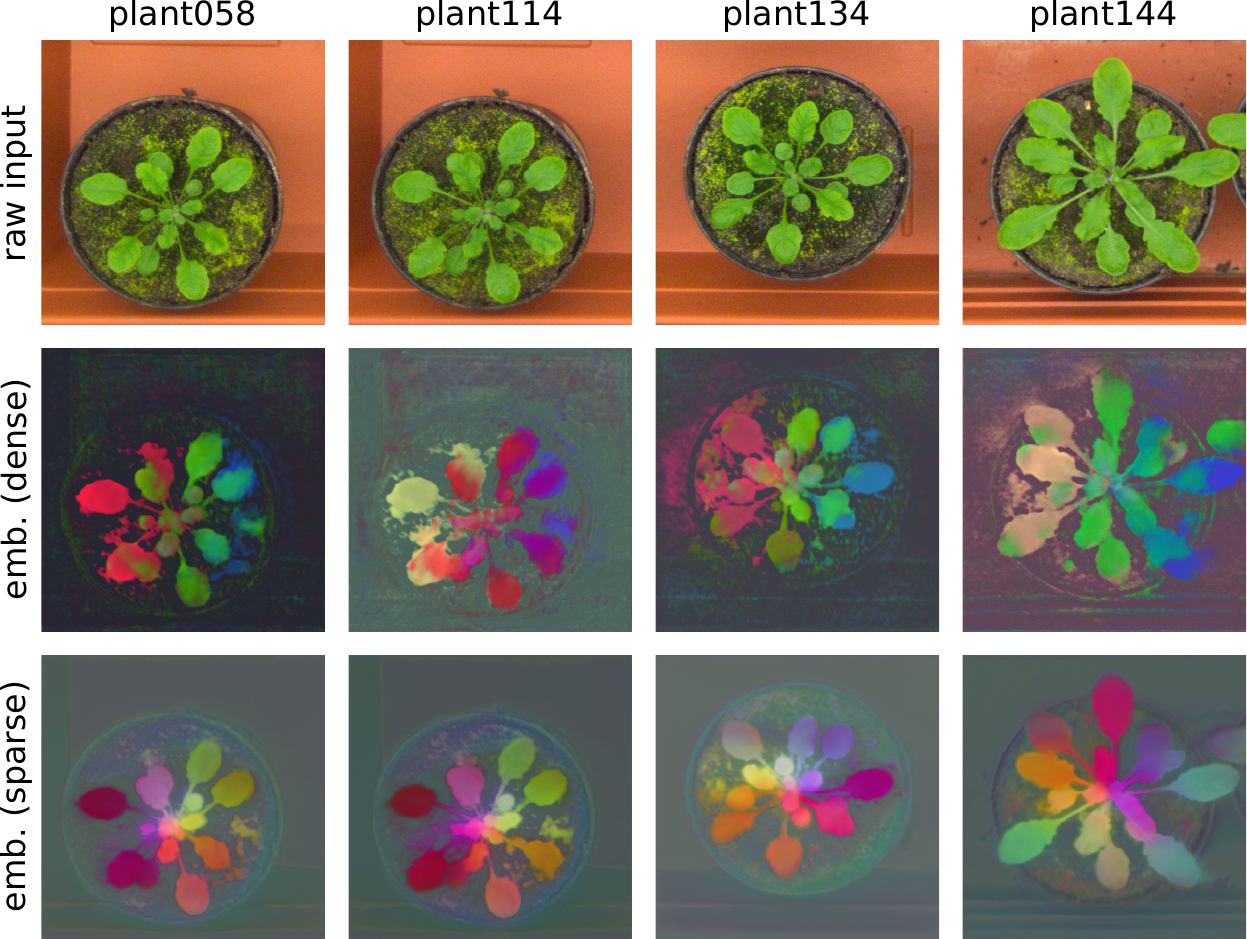}	
\end{flushleft}
\vspace{-0.5em}
\caption{(Top) Segmentation performance computed for the networks trained with limited annotation budget on the CVPPP validation set. Embeddings within the foreground semantic mask were clustered with mean-shift algorithm. (Bottom) Qualitative comparison of the embeddings trained in the \textit{dense} and \textit{sparse} setting. Four sample images where chosen from the CVPPP validation set.}
\label{tab:sparse_vs_dense}
\end{table}

\subsection{Momentum coefficient ($m$) exploration}
In this experiment, we explore the effect of the momentum coefficient $m$ used in the momentum update of the g-network parameters (see Sec.~\ref{ssec:sparse_single_obj}).
We use the train/val split of the CVPPP training set described in Sec.~\ref{ssec:g_net_ablation}.
Similar to \cite{He_2020_CVPR} we show in Table~\ref{tab:momentum_coeff} that the large momentum ($m=0.999$) performs best. We hypothesize that using slowly moving $g$ moving acts as a strong regularizer which prevents the embedding network $f$ to adapt too quickly to the spare ground truth signal.

\begin{table}[!htbp]

\begin{tabularx}{\columnwidth}{|>{\hsize=1.0\hsize\linewidth=\hsize}X|>{\hsize=0.6\hsize\linewidth=\hsize}X|>{\hsize=0.4\hsize\linewidth=\hsize}X|}
\hline
Momentum coefficient & SBD & $|DiC|$ \\
\hline
0.99 & 0.615 $\pm$ 0.042 & 5.2 $\pm$ 0.8 \\
0.995 & 0.622 $\pm$ 0.116 & 5.4 $\pm$ 0.7 \\
0.999 & \textbf{0.649 $\pm$ 0.045} & \textbf{4.5 $\pm$ 1.6} \\
\hline
\end{tabularx}
\vspace{.5em}
\begin{flushleft}
\vspace{-1em}
\includegraphics[width=.95\linewidth]{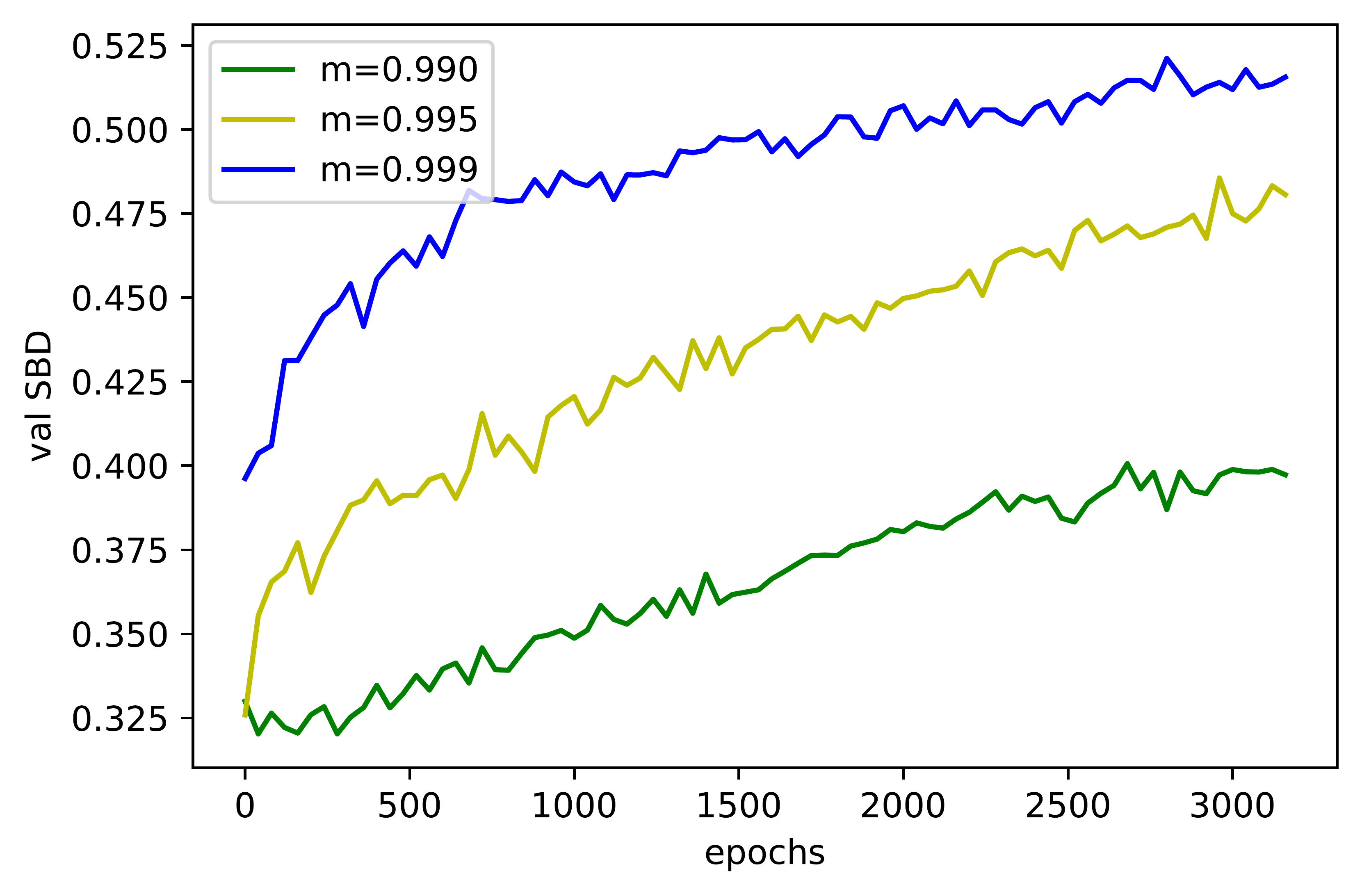}	
\end{flushleft}
\vspace{-1em}
\caption{The effect of the momentum coefficient value $m$ on the SPOCO performance. (\textbf{top}) Segmentation and counting scores. (\textbf{bottom}) Evolution of the validation score during training. SPOCO@01 was used for training.  Mean $\pm$ SD across 3 random samplings of the ground truth objects is shown.}
\label{tab:momentum_coeff}
\end{table}

\subsection{Kernel threshold ($t$) exploration}
\begin{figure}[!htbp]
\begin{center}
\includegraphics[width=1.0\linewidth]{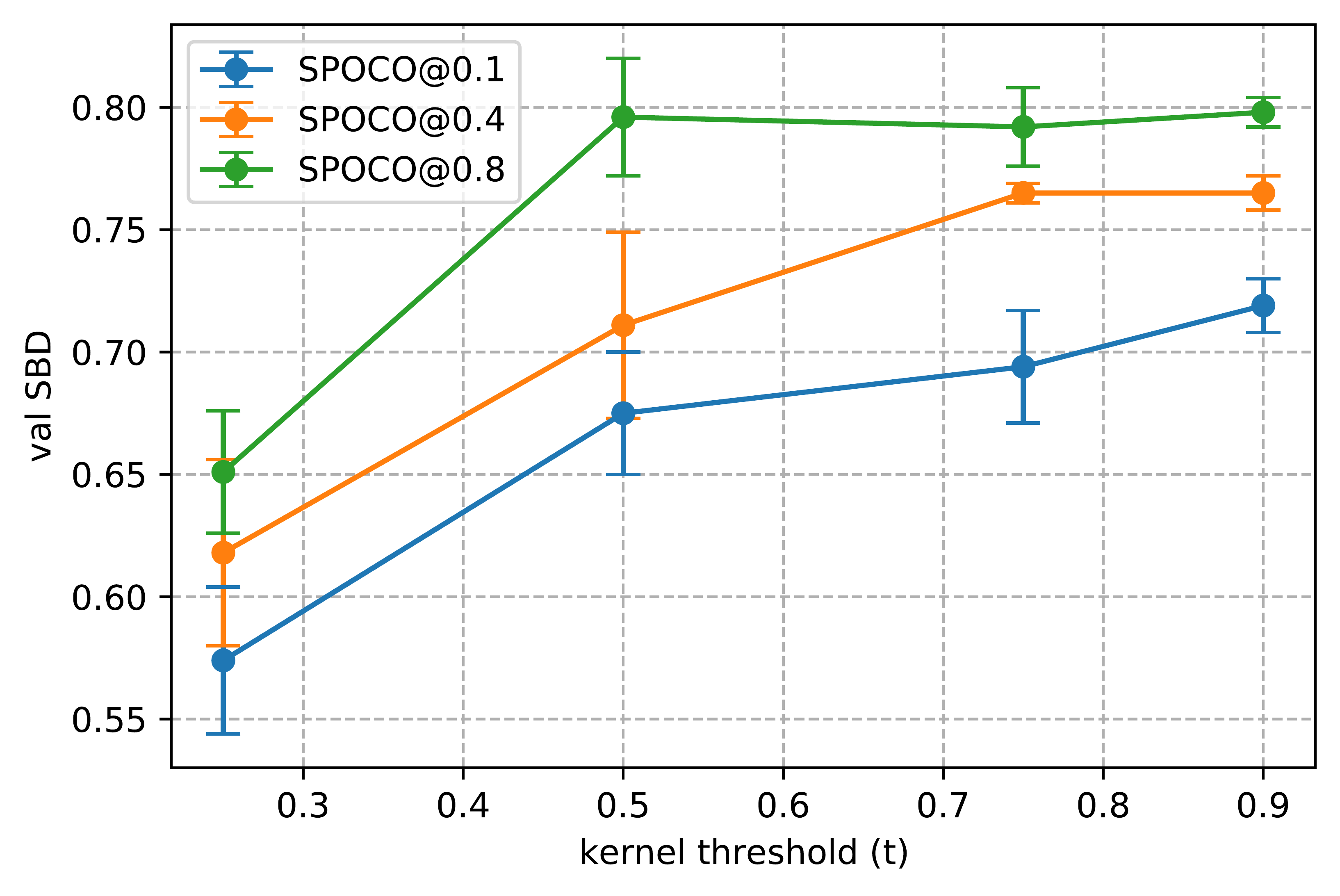}
\end{center}
\vspace{-1em}
   \caption{Effect of the kernel threshold $t$ on the segmentation performance at different ground truth objects sampling rates ($0.1, 0.4, 0.8$). SBD scores measured on the CVPPP validation set are shown for models trained with four values of $t$: $0.25, 0.5, 0.75, 0.9$. Mean $\pm$ SD are reported across 3 training runs for each (sampling rate, kernel threshold) pair. HDBSCAN ($min\_size=200$) is used for clustering.}
\label{fig:kernel_threshold} 
\end{figure}

Figure~\ref{fig:kernel_threshold} illustrates the effect of the kernel threshold parameter $t$ (Eq.~\ref{eq:object} in Sec.~\ref{ssec:single_obj}) on the SPOCO model performance. Choosing a large value (e.g. $t=0.9$) leads to a crisper, more separable embeddings than smaller values (e.g. $t \in \{0.25, 0.5, 0.75\}$). The difference in the final segmentation performance between a small and a large value of $t$ is especially apparent in the sparse annotation regime. Indeed, when training with only 10\%  (SPOCO@0.1) or 40\% (SPOCO@0.4) of ground truth objects, the mean SBD score improvement between $t=0.5$ and $t=0.9$ is $0.044$ and $0.054$ respectively. Although the performance gain is less pronounced when more supervision is provided (for SPOCO@0.8 the mean SBD reaches a plateau for $t \ge 0.5$), models trained with higher values of $t$ are more robust as shown by the low variance of the SBD score.\\
In our experiments, values of $t$ greater than $0.95$ lead to training instabilities.

\subsection{Cityscapes results}
In Tab~\ref{tab:one_vs_all} we compare two different training setups at different object sampling ratios for the Cityscapes dataset: (1) \textit{single-class} reported in the main text, where embedding network is trained separately on each semantic class and (2) \textit{class-agnostic} where all objects from all classes are used to train a single embedding network. The class-agnostic training works better for riders, cars, motorcycles and bicycles at all sampling levels. The single-class training is better for trucks, buses, and trains. We hypothesize that the class-agnostic setup learns better representation of objects from correlated classes (e.g. person and rider, motorcycle and bicycle), but it is detrimental to trucks, buses and trains due to heavy class imbalance. A per-class weighting of the instance-based term could be beneficial in the class-agnostic setting, which we leave for future work.

\begin{figure*}[b]
\begin{center}
\includegraphics[width=1.0\linewidth]{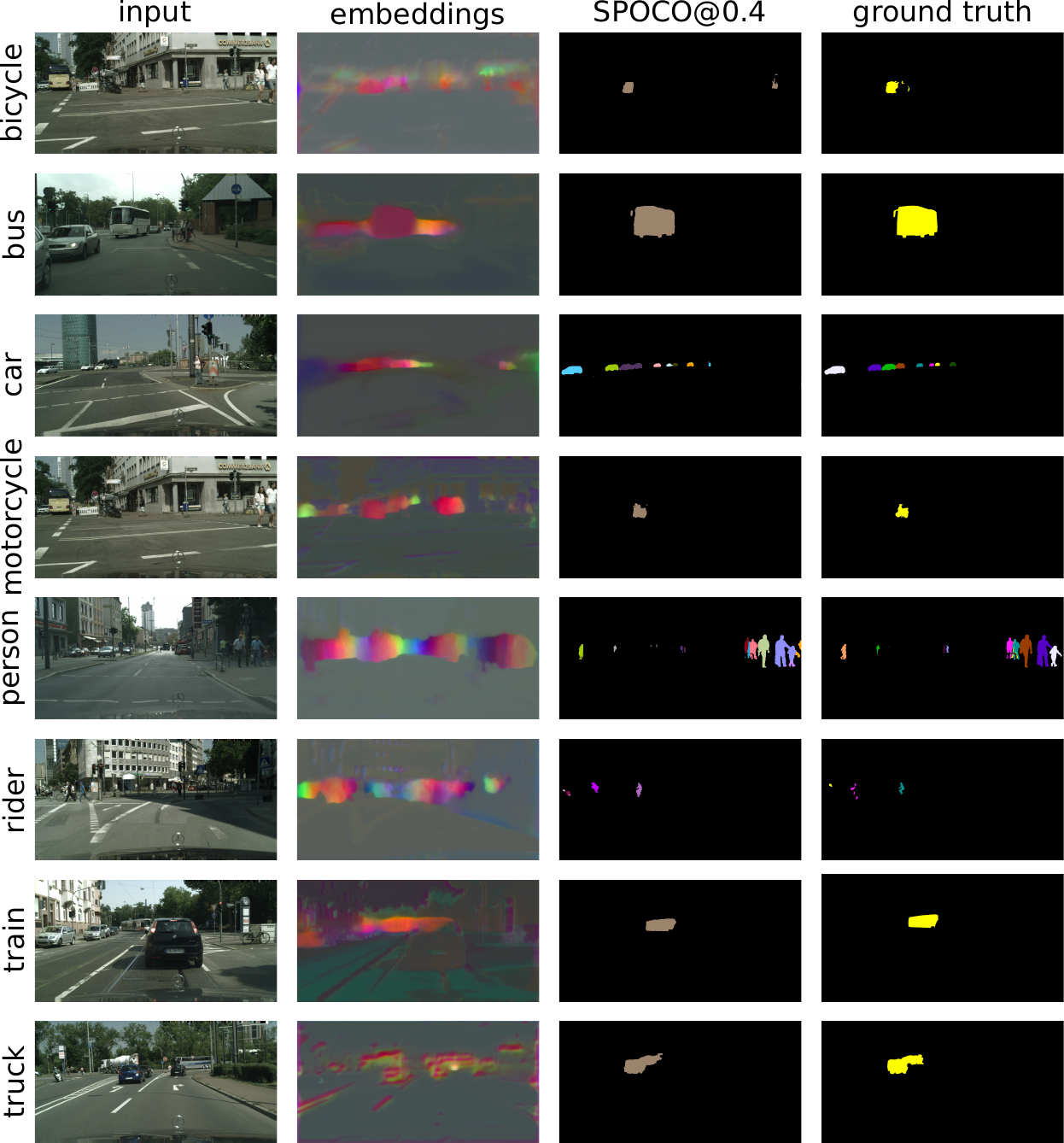}
\end{center}
\vspace{-.5em}
\caption{Qualitative results for different semantic classes on the Cityscapes validation set.}
\label{fig:cityscapes_qualitative}
\end{figure*}

Additional qualitative results for the weakly-supervised network (SPOCO@0.4) on the Cityscapes validation set can be found in Fig~\ref{fig:cityscapes_qualitative}. Apart from the segmentation results, we also show the embeddings leaned by the network trained on objects from a given semantic class.
For underrepresented classes such as motorcycle, train and truck the final segmentation strongly relies on the segmentation mask given by the pre-trained semantic segmentation model (DeepLabV3 \cite{deeplabv3_2018})

\newgeometry{left=0.8125in,top=0.9625in,bottom=0.9625in}
\begin{table*}[t]
\begin{center}
\begin{tabular}{|c|c|c|c|c|c|c|c|c|c|}
\hline
Method  & person & rider & car & truck & bus & train & motorcycle & bicycle & \textbf{average} \\
\hline
single-class@0.1  & 0.190 & 0.360 & 0.236 & \textbf{0.438} & \textbf{0.481} & \textbf{0.490} & 0.424 & 0.204 & \textbf{0.353} \\
class-agnostic@0.1   & \textbf{0.197} & \textbf{0.430} & \textbf{0.282} & 0.243 & 0.276 & 0.167 & \textbf{0.468} & \textbf{0.261} & 0.291 \\
\hline
single-class@0.4  & \textbf{0.230} & 0.396 & 0.301 & \textbf{0.558} & \textbf{0.601} & \textbf{0.594} & 0.405 & 0.214 & \textbf{0.412} \\
class-agnostic@0.4   & 0.207 & \textbf{0.459} & \textbf{0.332} & 0.260 & 0.336 & 0.223 & \textbf{0.471} & \textbf{0.266} & 0.319 \\
\hline
single-class@1.0  & \textbf{0.260} & 0.451 & 0.331 & \textbf{0.604} & \textbf{0.637} & \textbf{0.656} & 0.464 & 0.266 & \textbf{0.459} \\
class-agnostic@1.0   & 0.259 & \textbf{0.463} & \textbf{0.410} & 0.370 & 0.395 & 0.378 & \textbf{0.478} & \textbf{0.296} & 0.381 \\
\hline
\end{tabular}
\end{center}
\vspace{-.3em}
\caption{Comparison of SPOCO trained in a single-class vs class-agnostic settings at different sampling ratios. Shown are mAP@0.5 scores computed on the Cityscapes validation set.}
\vspace{-.3em}
\label{tab:one_vs_all}
\end{table*}

\restoregeometry

{\small
\bibliographystyle{ieee}
\bibliography{spoco}
}

\end{document}